\documentclass[12pt]{iopart}

\usepackage{graphicx}%
\usepackage{multirow}%
\expandafter\let\csname equation*\endcsname\relax

\expandafter\let\csname endequation*\endcsname\relax

\usepackage{amsmath, amssymb}
\usepackage[title]{appendix}%
\usepackage{xcolor}%
\usepackage{textcomp}%
\usepackage{manyfoot}%
\usepackage{booktabs}%
\usepackage{algorithm}%
\usepackage{algorithmicx}%
\usepackage{algpseudocode}%
\usepackage{listings}%

\begin{document}

\title[]{Continual Learning with Hebbian Plasticity in Sparse and Predictive Coding Networks: A Survey and Perspective}

\author{Ali Safa} 

\address{College of Science and Engineering, Hamad Bin Khalifa University, Doha, Qatar} 
\ead{asafa@hbku.edu.qa} 
\vspace{10pt}
\begin{indented}
\item[]November 2024
\end{indented}

\begin{abstract}
Recently, the use of bio-inspired learning techniques such as Hebbian learning and its closely-related Spike-Timing-Dependent Plasticity (STDP) variant have drawn significant attention for the design of compute-efficient AI systems that can continuously learn on-line at the edge. A key differentiating factor regarding this emerging class of neuromorphic continual learning system lies in the fact that learning must be carried using a data stream received in its natural order, as opposed to conventional gradient-based offline training, where a static training dataset is assumed available a priori and randomly shuffled to make the training set independent and identically distributed (i.i.d). In contrast, the emerging class of neuromorphic continual learning systems covered in this survey must learn to integrate new information on the fly in a non-i.i.d manner, which makes these systems subject to catastrophic forgetting. In order to build the next generation of neuromorphic AI systems that can continuously learn at the edge, a growing number of research groups are studying the use of Sparse and Predictive Coding-based Hebbian neural network architectures and the related Spiking Neural Networks (SNNs) equipped with STDP learning. However, since this research field is still emerging, there is a need for providing a holistic view of the different approaches proposed in the literature so far. To this end, this survey covers a number of recent works in the field of neuromorphic continual learning based on state-of-the-art Sparse and Predictive Coding technology; provides background theory to help interested researchers quickly learn the key concepts; and discusses important future research questions in light of the different works covered in this paper. It is hoped that this survey will contribute towards future research in the field of neuromorphic continual learning.
\end{abstract}

%
%
%
%
%
\section*{Supplementary Material}
Please find the code for a demonstration of Continual Learning with Sparse Coding Networks (presented in Section \ref{expecl}) at \texttt{https://tinyurl.com/y8n5a3tf}

\section{Introduction}

The remarkable ability of biological systems to learn and adapt in real-time to their environment while consuming little power compared to conventional Deep Learning systems has been a major motivation behind the study of neuromorphic computing \cite{cath1, cath2, surv1, surv2}. Indeed, biological entities significantly differ from the current Deep Learning AI systems in terms of their \textit{learning paradigm} \cite{khacefstdp}. While the large majority of today's AI systems rely on the availability of training datasets \textit{captured beforehand}, biological entities are capable to learn on the fly using the perception data (e.g., vision) received in a real-time order \cite{asafapeople}. Hence, biological entities seem to be capable of learning in non-stationary data streams where data points cannot be assumed independent and identically distributed (non-i.i.d) \cite{9349197, asafaslam2}. This learning paradigm is referred to as \textit{Continual Learning} in literature \cite{9349197}. This comes in great contrast with the gradient-based \textit{offline} training setting, where the offline training data is randomly shuffled in order to make the data source i.i.d, which is needed for stochastic gradient descent to converge successfully \cite{9349197}. 
 
In recent years, the remarkable ability of biological entities for continual learning has triggered research in the emerging field of \textit{neuromorphic} continual learning, which seeks to study how bio-inspired neural network architectures such as non-spiking Artificial Neural Networks (ANNs) equipped with Hebbian learning and Spiking Neural Networks (SNNs) equipped with Spike-Timing-Dependent Plasticity (STDP) can be used for designing continual learning systems, without suffering from the compute- and memory-intensive overheads of traditional Deep Learning systems \cite{ororbia1, annabi, ororbia3, asafaslam1, antonov, alfredo, chip1}. Following these realisations, a number of works covering a wide range of applications have been proposed, from task- and class-incremental learning setups \cite{ororbia1} to the continual learning of object detection and robot navigation \cite{asafapeople, asafaslam2}.

Studying the conception of neuromorphic continual learning system is an important emerging topic as it has the potential to enable a new class of extreme edge devices that can learn while performing inference at the edge \cite{stdpalijon}. This novel mode of operation is significantly different to the current way that most neuromorphic systems operate with weights learned during an \textit{offline} training phase that do not change or adapt during the inference phase \cite{surv2}. Hence, the study of neuromorphic continual learning will lead to novel functionalities and features such as:
\begin{itemize}
    \item The \textit{personalization} of extreme edge devices to their specific environment or user, which will lead to better algorithmic performance \cite{asafaslam2}.
    \item Alleviating privacy issues: these new algorithms will be able to refine the weights of their neural networks without resorting to compute-expensive offline training phases, which require a transfer of data from the extreme edge device to the cloud \cite{stdpalijon}.
    \item On-chip learning: understanding how to design learning algorithms that can learn on a stream of real-world data shown in its real-time order considerably reduces the memory requirements compared to conventional shuffled datasets used in traditional offline training. This reduction in memory requirements would in turn reduce the chip area and power consumption of the overall integrated circuits \cite{surv3}.

\end{itemize}

But, to the best of our knowledge, a holistic survey giving an ensemble view of the different state-of-the-art works carried in the field of \textit{neuromorphic continual learning} has not yet been provided, surely due to the more recent nature of this field compared to other sub-fields of neuromorphic computing which have been researched for a longer time. Hence, inspired by a number of prior surveys \cite{cath1, cath2, surv1, surv2} covering neuromorphic topics such as local synaptic plasticity \cite{khacefstdp} and event-based computation \cite{stdpalijon}, this paper provides a holistic survey of recent works in the field of neuromorphic continual learning, using state-of-the-art Sparse and Predictive Coding architectures which have gained huge interest in this field. Compared to the closely-related survey proposed by R. Mishra \textit{et al.} \cite{surv3}, this paper provides a \textit{complementary} view of works in the field of neuromorphic continual learning, where focus is mostly made on describing \textit{system-level} works reported in literature, complementary to the focus on \textit{specific techniques} mostly adopted in the survey of R. Mishra \textit{et al.} \cite{surv3}. 

In this survey, both Hebbian learning in non-spiking ANNs and its \textit{closely-related} STDP variant in SNNs are covered, since it is widely accepted that STDP can be seen as a specific implementation of Hebbian learning in the spiking domain \cite{stdphebbi}, which motivates our covering of both related notions in this paper. Since many of the related works are based on the theoretical concepts of Sparse Coding and Predictive Coding \cite{fristonpredic, 9892362}, this paper also provides useful background theory on the key concepts behind the neuromorphic continual learning systems reported in literature. Finally, this survey discusses the future perspectives and suggests a number of future research questions for further advancing the field of neuromorphic continual learning.




This paper is organized as follows. Preliminaries on the relation between Continual Learning and Sparse \& Predictive Coding Networks are provided in Section \ref{prelim}. Background theory is provided in Section \ref{background}. An overview of notable works in continual learning with non-spiking Hebbian plasticity is covered in Section \ref{clhebbian}. Transitioning to the spiking domain, an overview of notable works in continual learning with SNNs and STDP plasticity is given in Section \ref{clstdp}. Discussions and future perspective on bio-plausible continual learning techniques are provided in Section \ref{future}. Finally, conclusions are provided in Section \ref{conclusion}.

\section{Preliminaries on the relation between Continual Learning and Sparse \& Predictive Coding Networks}
\label{prelim}

\subsection{Motivations}
\label{motiv}

The approach adopted in this survey seeks to cover the use of Sparse \& Predictive Coding Neural Networks for Continual Learning applications, as state-of-the-art neural network design approaches that closely incorporate bio-plausible Hebbian learning \cite{lin2018sparse}. Following background literature, we use the terminology of \textit{Predictive} Coding to refer to a multi-layer cascade of single Sparse Coding layers \cite{fristonpredic, 9892362}. 

It is important to state that this survey will distinguish the covered works between the non-spiking and spiking cases. In the non-spiking case, the neural architectures correspond to a connectionist implementation of Sparse or Predictive Coding algorithms. In the spiking case on the other hand, the neural architectures make use of spiking neurons and represent a Spiking Neural Network (SNN) implementation of Sparse or Predictive Coding algorithms. In addition, the SNN cases make use of the Spike-Timing-Dependent Plasticity (STDP) rule \cite{stdphebbi}, as a spiking implementation of Hebbian learning. We make this distinction between the non-spiking and spiking case to be in-line with the hardware-efficient SNN-STDP processors that have been remarkably designed by the Neuromorphic community \cite{cath2,odin}. 

\subsubsection{On the link between Sparse \& Predictive Coding networks and Hebbian learning}

Sparse and Predictive Coding architectures are known to be tightly linked with the concept of Hebbian learning, since it can be mathematically shown that the learning that takes place within Sparse or Predictive Coding systems when implemented as neural networks boils down to local Hebbian learning mechanisms \cite{lin2018sparse, neuralcodeframe} (see Equation \ref{hebbianrule} in Section \ref{sconly} of this survey). Today, a growing research community is interested in studying Hebbian learning networks via the use of Sparse and Predictive Coding architectures, since this setting provides a clear mathematical model for understanding what precise objective function the Hebbian plasticity mechanisms are optimizing \cite{surv3, annabi, asafaslam1, lin2018sparse, neuralcodeframe}. This is in contrast to the use of prior-art Hebbian network architectures such as Reservoir-style and Winner-Take-All (WTA) networks where the energy function minimized by the Hebbian network is not clearly formulated, leading to un-modeled and less-predictive behaviors \cite{9892362}. On the other hand, this key observation made by a number of researchers in the Neuromorphic community about the tight link between the concise mathematical formalism of Sparse or Predictive Coding theory and Hebbian learning has enabled the inception of more powerful Hebbian learning networks, enabling the tackling of new challenging tasks such as drone navigation, people detection, agent control and so on \cite{asafapeople,ororbia3, asafaslam1}. 

\subsubsection{On the link between Continual Learning and Sparse \& Predictive Coding networks}

Sparse and Predictive Coding networks (which implement local Hebbian-style plasticity) are increasingly being studied under a Continual Learning setting because they are believed to be key enablers of future Neuromorphic devices that could personalize themselves to their working environment and learn on the fly in an energy-efficient way (like biological entities, but unlike current offline-trained deep learning systems). In Deep Learning approaches, Continual Learning is often achieved by computing complex regularization terms involving second-order Hessian matrix computations and their related variants \cite{EWC}. But another subcategory of Continual Learning techniques within Deep Learning are referred to as \textit{Parameter Isolation} methods \cite{9349197}. As their name suggest, Parameter Isolation methods seek to perform continual learning by allocating (in a hard or soft manner) portions of the total neural network resources to each different latent task domain that the network is seeking to learn. This corresponds to what Sparse and Predictive Coding systems can achieve due to their sparsity constraints and re-projection error minimization during learning. Indeed, the sparsity constraints in their neural activations leads to a soft compartmentalizing and isolation of their weights, since output neurons that are not firing do not provoke a change in their associated synapse strength, by virtue of Hebbian Learning. On top of this, the re-projection error minimization also helps with the compartmentalizing of the learned information since it encourages the weights to learn complementary visual fields (i.e., a diverse set of matched filters) where each visual field is better tailored to a particular input data context within the stream of non-identically-distributed data used during the continual learning process. In Section \ref{expecl} we provide an illustrative experiment clearly showing the usefulness of Sparse and Predictive Coding networks for Continual Learning.

\subsubsection{On the importance of Sparse \& Predictive Coding networks, Hebbian learning and Continual Learning}

Finally, we would like to briefly motivate and explain why studying the application of Sparse and Predictive Coding networks (which implement Hebbian plasticity) for Continual Learning is important from both a computational Neuroscience and Engineering point of view. 

First, from a computational Neuroscience point of view, it is well known that backpropagation-style learning algorithms are, in the overwhelming majority of works, not considered biologically plausible due to their non-local nature \cite{backpropnot}. Now, when it comes to Continual Learning, using regularization-based techniques that compute variants of Hessian matrices for the network weights \cite{EWC} make this even more implausible from a biological point of view. On the other hand, it is well-known that Hebbian learning has strong biological plausibility \cite{Bi10464}, and the same can be said about Sparse and Predictive Coding Networks which have been proposed in a number of past seminal works as plausible architecture for e.g., the V1 region behavior in the brain \cite{olhausend}. In addition to this, it is well-known that biological agents have a remarkable capability at adapting their behavior and \textit{learning in real time}, as in the Continual Learning case. Hence, this leads to a key research question: how can we design Continual Learning systems using more bio-plausible architectures such as Sparse and Predictive Coding networks (which learn though local Hebbian plasticity). 

Second, from an engineering point of view, it is believed by the community that being able to build energy-efficient continual learning systems that can learn and personalize themselves on the fly would constitute a major realization in the Neuromorphic Computing field, compared to the ever dominance of offline training algorithms in today’s deep learning systems \cite{surv3}. Hence, performing research in this field has a huge potential since it could potentially enable the creation of a whole range of novel IoT and edge consumer devices that could intimately adapt themselves to their user or environment, so as to attain higher performances, without resorting to costly offline training and its privacy preservation concerns. This constitutes a major motivation behind the study of Hebbian-driven Sparse and Predictive Coding Networks for continual learning, as an emerging, state-of-the-art approach for designing local learning algorithms. These local Hebbian learning networks could then be fit for deployment in Hardware-efficient Neuromorphic chips developed by the vibrant Neuromorphic hardware community \cite{cath2,odin}.

\subsection{Experimental motivation on the usefulness of Sparse \& Predictive Coding networks for Continual Learning}
\label{expecl}

In addition to the preliminary motivations introduced in Section \ref{motiv} on the tight links between Sparse \& Predictive Coding networks, Hebbian learning and Continual Learning, this section provides an introductory experimental illustration of \textit{the usefulness of Sparse \& Predictive Coding networks for tackling Continual Learning scenarios}.
\begin{figure}
\centering
\includegraphics[scale = 0.6]{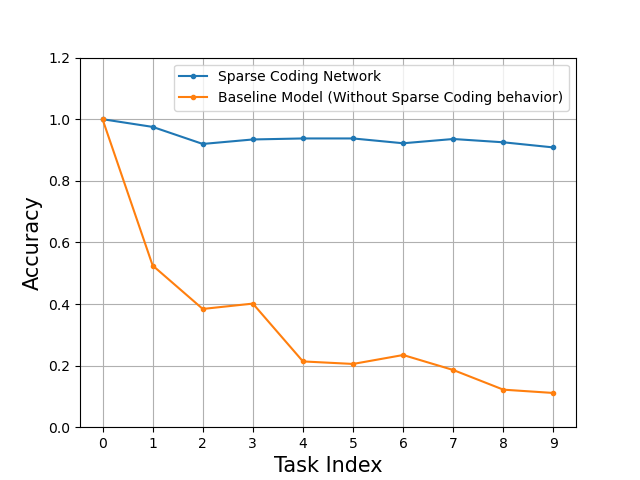}
\caption{\textit{\textbf{Illustration of the usefulness of Sparse Coding Networks for Continual Learning applications.} Compared to the baseline model (supervised linear 1-layer network), the 1-layer Sparse Coding network outperforms the baseline model in terms of classification accuracy retention, as new digit classes are fed to the networks during the continual learning process.}} 
\label{normalvssc}
\end{figure}

In this continual learning experiment, we compare the classification accuracy of a 1-layer Sparse Coding network to the accuracy of a fully-connected 1-layer baseline model (i.e., without Sparse Coding behavior). We implement both networks in \texttt{python} using the \texttt{pytorch} library \cite{pytorch}. Classification is done using population coding \cite{popcode}. Our continual learning scenario is set up as follows. Images from the \texttt{digits} dataset of the \texttt{sklearn} library \cite{sklearn} are fed in a class-incremental manner to each neural network model. This means that all images corresponding to the class "0" are first used for learning; then, images of the class "0" are not shown anymore and only images from the class "1" are shown. The continual learning process continues until all the 10 classes are covered. This continual learning setup is naturally challenging since it makes the learning prone to \textit{catastrophic forgetting}. Indeed, when feeding the images of class "1", the models could be prone to over-fitting on the images of class "1" and in turn, forget what was learned with the images of class "0" etc.

Fig. \ref{normalvssc} shows the result of this illustrative experiment, where it can be clearly remarked that the Sparse Coding network significantly outperforms the baseline model in terms of accuracy retention as new classes are fed to the networks. Hence, this experiment clearly motivates the use of Sparse \& Predictive Coding networks for continual learning applications and explains why a growing number of research teams are now exploring the use of Sparse Coding-type networks for building the next generation of Neuromorphic continual learning systems. 

Finally, the source code for the experiment in Fig. \ref{normalvssc} is released as open-source at \texttt{https://tinyurl.com/y8n5a3tf}

\section{Background theory}
\label{background}

This Section provides useful background information on the main neural network components used in the various works that are surveyed in this paper. First in the \textit{non-spiking} domain, since many of the neural network designs surveyed in this paper are based on the Sparse Coding (SC) \cite{olhausend} and related Predictive Coding architectures (PC) \cite{predcode2, predcode}, Section \ref{scpc} overviews the working principle behind SC and PC, as well as providing a view of how Hebbian learning techniques are used within the SC and PC contexts. Then, transitioning to the \textit{spiking} case, Section \ref{snnstdp} discusses how SNN architectures based on the SC principle are built in literature, and provides a view of how STDP learning techniques are used within the SNN context. Finally, Section \ref{clproblem} introduces the Continual Learning (CL) problem and its effect on learning performance.


\subsection{Sparse Coding and Predictive Coding Neural Architectures with Hebbian Learning}
\label{scpc}

\subsubsection{Sparse Coding with ANNs and Hebbian plasticity}
\label{sconly}
\hfill \break

An important number of state-of-the-art works in literature model the neural activity of biological agents through Sparse Coding (SC) and Predictive Coding (PC) neural networks \cite{fristonpredic}, \cite{9892362}. This is motivated by the fact that SC and PC have been shown to be bio-plausible models of cortical neural ensembles, as observed in the mushroom body of an insect's brain \cite{liang2021can}. In a probabilistic sense, SC models the output and weight distribution of ANNs equipped with Hebbian learning through an identically-distributed Gaussian likelihood model with a Laplacian prior on the neural activity $c$:
\begin{equation}
    p(c|o,\Phi) \sim \exp{(-|| \Phi c - o ||_2^2)} \exp{(-\lambda ||c||_1)}
    \label{sparsemodel}
\end{equation}
where $o$ is the input of dimension $N$, $c$ is the output of dimension $M$, $\Phi$ is the $N \times M$ weight matrix of the layer (also called \textit{dictionary}), and $\lambda$ is a hyper-parameter setting the scale of the Laplacian prior. Using a Laplacian prior is motivated by the fact that its distribution shape promotes sparsity in the output neural activation, in an analogous way to how bio-plausible SNNs using e.g., Leaky Integrate-and-Fire (LIF) neurons promote sparsity in their output codes \cite{9892362}.  

In SC (\ref{sparsemodel}), the inference of the output activation $c$ and the learning of the network weights $\Phi$ can be obtained through \cite{9892362}:
\begin{equation}
    C, \Phi = \arg \min_{C, \Phi} \sum_l  || \Phi c_l - o_l ||_2^2 + \lambda ||c_l||_1 \textbf{ with } C = \{c_l, \forall l\}
    \label{dlbp}
\end{equation}
Equation \ref{dlbp} is typically solved via Proximal Stochastic Gradient Descent \cite{NEURIPS2019d073bb8d}, by \textit{alternating} between: \textit{a)} the inference of $c_l$, given the current input $o_l$ and the weight $\Phi$ and \textit{b)} the learning of $\Phi$, given the current $c_l$ and $o_l$. 

Neural network layers solving the SC problem (\ref{dlbp}) can be implemented following the dynamical system given in (\ref{dynamics}), where $T$ denotes the transpose, $\eta_c$ is the coding rate, $\eta_d$ is the learning rate and $\textbf{Prox}_{\lambda ||.||_1}$ is the proximal operator to the $l_1$ norm (non-linearity) \cite{lin2018sparse}. For each input $o_i$, the neural and weight dynamics of the Hebbian network follows the update rules in (\ref{dynamics}) for an arbitrary number of iterations $N_{it}$, in order to infer the corresponding $c_l$ and learn $\Phi$ \cite{9892362}. 
\begin{equation}
     \begin{cases}
     c_l \leftarrow \textbf{Prox}_{\lambda ||.||_1} \{ c_l - \eta_c\Phi^T (\Phi c_l - o_l)  \} \\ 
     \Phi \leftarrow \Phi - \eta_d (\Phi c_l - o_l) c_l^T
    \end{cases}   
    \label{dynamics}
\end{equation}
with $\textbf{Prox}_{\lambda ||.||_1}$ acting as the neural non-linearity:
\begin{equation}
\textbf{Prox}_{\lambda ||.||_1}(c_i) = \text{sign}(c_i) \max(0, |c_i| - \eta_c \lambda), \forall i 
    \label{proxop}
\end{equation}

\begin{figure}
\centering
\includegraphics[scale = 0.56]{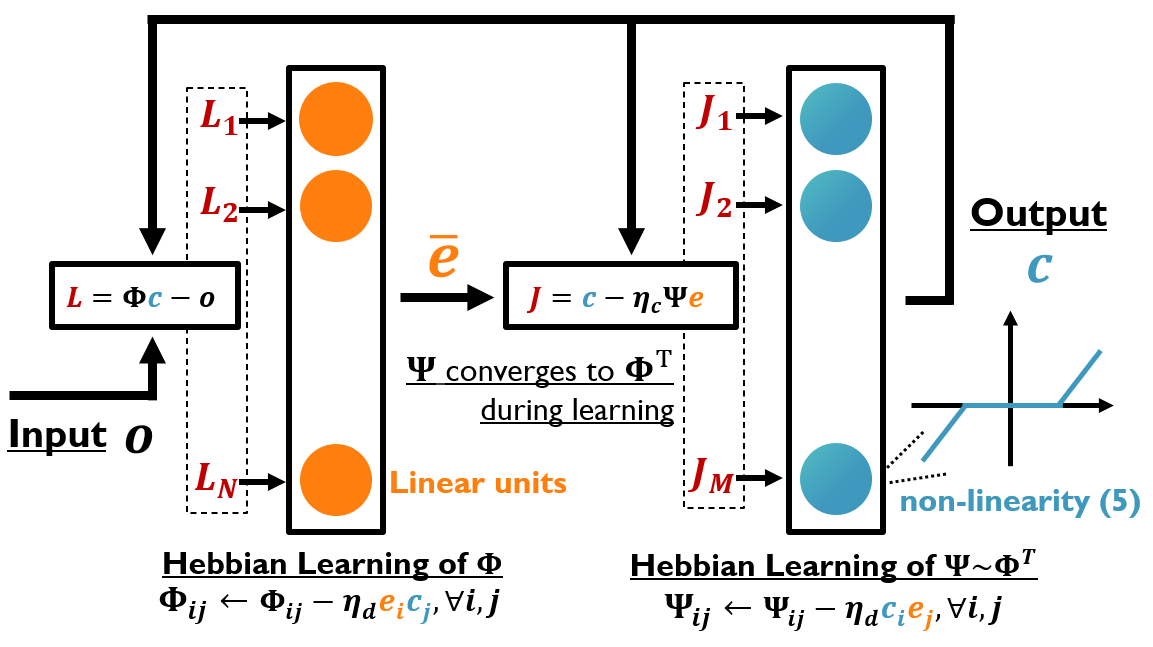}
\caption{\textit{\textbf{Baseline Hebbian network architecture used in this work.} The dynamics of the network follow (\ref{dynamics}) and minimize (\ref{dlbp}), given subsequent input vectors $o$. Each layer possesses its own weight matrix $\Phi, \Psi$ which evolve through Hebbian plasticity ($\Psi \sim \Phi^T$ in (\ref{dynamics}), as an \textit{independent}, local set of weights).}} 
\label{basenetarch}
\end{figure}

Algorithm \ref{pursuitalg} summarizes the alternating-descent procedure classically used to solve (\ref{dlbp}) \cite{9892362}. This procedure will also be used by the neural network architecture shown in Fig. \ref{basenetarch}, essentially executing Algorithm \ref{pursuitalg} through its explicit neural wiring and Hebbian plasticity. 
 \begin{algorithm}
 \caption{joint Dictionary Learning and Basis Pursuit}
 \label{pursuitalg}
 \begin{algorithmic}[1]
 \renewcommand{\algorithmicrequire}{\textbf{Input:}}
 \renewcommand{\algorithmicensure}{\textbf{Output:}}
 \Require $\bar{s}_k \hspace{3pt} \forall k$: input vector stream, $\eta_c$: coding rate, $\eta_d$: learning rate, $\lambda_1$: regularization, $N_{it}$: number of iterations.  
 \\ \textbf{Init.}: $\Phi_{ij} \xleftarrow{} \mathcal{N}(0,\sigma_w)$ (zero-mean normal distribution with std. deviation $\sigma_w \sim 0.01$), $\bar{c} \xleftarrow{} 0$
  \For {$l \in \{1,...,T_{end} \}$ (feed data sequence)}
  \For{$p = 1,...,N_{it}$ (local coding iterations)}
  \State $\bar{c}_l \xleftarrow{} \textbf{Prox}_{\eta_c \lambda_1 ||.||_1} \{\bar{c}_l - \eta_c \Phi^T(\Phi \bar{c}_l - \bar{s}_l) \}$ // see (\ref{dynamics})
  \EndFor
  \For{$p = 1,...,N_{it}$ (local learning iterations)} 
  \State $\Phi \xleftarrow{} \Phi - \eta_d(\Phi \bar{c}_l - \bar{s}_l) \bar{c}^T_l$ 
  \EndFor
  \EndFor
 \end{algorithmic} 
 \end{algorithm}

From a neural network topology point of view, the dynamical system of (\ref{dynamics}) can be implemented as the network architecture in Fig. \ref{basenetarch}, composed of an \textit{output} layer using the non-linearity (\ref{proxop}) and a \textit{linear} layer estimating the error vector $\Bar{e}$, where all weight updates follow the form of the standard Hebbian rule \cite{9892362}:
\begin{equation}
     w_{ij} \xleftarrow[]{} w_{ij} - \eta_d y_i x_j
     \label{hebbianrule}
\end{equation}

In (\ref{hebbianrule}), $\eta_d$ is the learning rate, $y_i$ is the post-synaptic (or output) activation value of neuron $i$ and $x_j$ is the pre-synaptic (or input) value received by the $j^{th}$ synapse of neuron $i$. The error vector $e$ outputted by the linear layer in Fig. \ref{basenetarch} encodes the re-projection error in (\ref{dlbp}) as:
\begin{equation}
    e = \Phi c - o
    \label{erroreq}
\end{equation}

In Fig. \ref{basenetarch}, Hebbian learning (\ref{hebbianrule}) is applied to the weight matrix $\Phi$ of the linear layer following:
\begin{equation}
     \Phi_{ij} \xleftarrow[]{} \Phi_{ij} - \eta_d e_i c_j
     \label{hebbianrule1}
\end{equation}
where pre-synaptic $c_j$ is the $j^{th}$ element of the output layer and post-synaptic $e_i$ is the $i^{th}$ element of the linear layer with output vector $e$ (\ref{erroreq}).

Similarly, Hebbian learning is locally applied to the weight matrix $\Psi$ of the output layer:
\begin{equation}
     \Psi_{ij} \xleftarrow[]{} \Psi_{ij} - \eta_d c_i e_j
     \label{hebbianrule2}
\end{equation}
making $\Psi$ converge to the \textit{transposed} of the dictionary weight matrix $\Phi$ during the learning process $\Psi \sim \Phi^T$ (since the pre-synaptic and post-synaptic signals are swapped between $\Psi$ and $\Phi$) \cite{9892362}.

\subsubsection{Predictive Coding with ANNs and Hebbian platicity}
\label{predictivecode}
\hfill \break

Closely related to the SC problem, Predictive Coding (PC) can be seen as a multi-layer SC system where each layer $r$ not only minimizes its own re-projection error $||\Phi_r c_r - o_r ||_2^2$ between its output sparse code $c_r$ and input observation $o_r$, but also seeks to minimize the re-projection error $||\Phi_{r+1} c_{r+1} - o_{r+1}||_2^2$ of the next downstream SC layer $r+1$, with $o_{r+1}=c_{r}$. Hence, each SC layer is also trying to \textit{predict} the error done by the layer coming after it, in order to minimize the global loss function of the complete system, corresponding to the sum of all re-projection errors \cite{predictivecoding}:
\begin{equation}
    \Phi_{1,...,R}, c_{1,...,R} = \arg \min_{\Phi_{1,...,R}, c_{1,...,R}} \sum_{r = 1}^{R} ||\Phi_r c_r - c_{r-1} ||_2^2 + \lambda ||c_r||_1
    \label{predictivecodingg}
\end{equation}
where $R$ denotes the number of cascaded SC layers and $c_0=o$ is the input observation data. The PC system can be implemented as a cascade of SC layers (see Fig. \ref{basenetarch}), where each layer $r$ also receives \textit{top-down} connection $v_{r}=\Phi_{r+1}c_{r+1}$ from the output sparse code $c_{r+1}$ of the next layer $r+1$, through its weight matrix $\Phi_{r+1}$. Thanks to the availability of this top-down connection $v_r$, each SC layer $r$ can also correct for the re-projection error of the next layer $r+1$ by upgrading the iterative process (\ref{dynamics}) to the following \cite{predictivecoding}:
\begin{equation}
     \begin{cases}
     c_r \leftarrow \textbf{Prox}_{\lambda ||.||_1} \{ c_r - \eta_c\Phi_r^T (\Phi_r c_r - c_{r-1}) - \eta_c (v_r - c_r)\} \\ 
     \Phi_r \leftarrow \Phi_r - \eta_d (\Phi_r c_r - c_{r-1}) c_r^T
    \end{cases}   
    \label{dynamicsPC}
\end{equation}
where the new term $(v_r - c_l)$ corresponds to the gradient of $||\Phi_{r+1}c_{r+1} - c_r||_2^2$ in function of $c_r$.
\begin{figure}
\centering
\includegraphics[scale = 0.56]{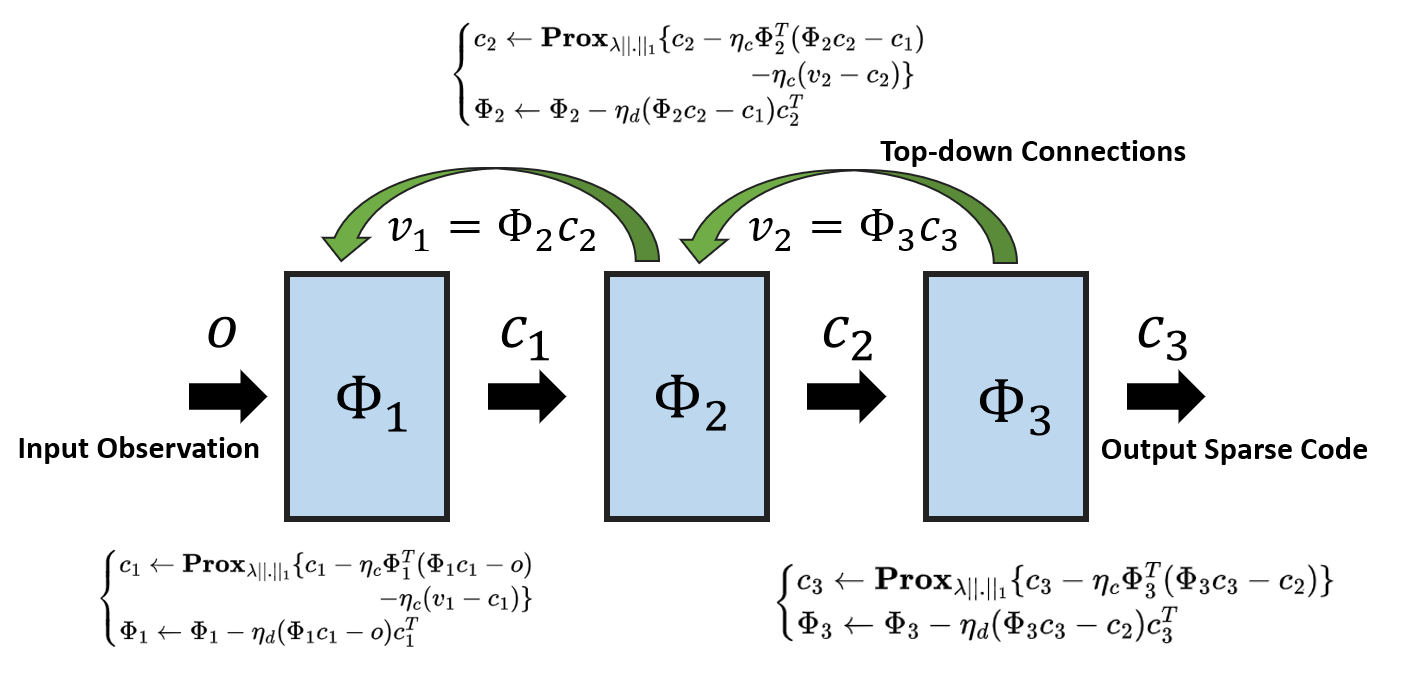}
\caption{\textit{\textbf{Predictive Coding neural network architecture.} In this example, a 3-layer PC network is shown, where each blue box represents a SC network instantiation of the Hebbian network shown in Fig. \ref{basenetarch}. The arrows indicate the top-down connections from each upstream layer to its previous one. }} 
\label{predictivecodeimg}
\end{figure}
Therefore, by cascading an arbitrary number $R$ of SC layers following Fig. \ref{predictivecodeimg}, and by applying Hebbian learning to the weights as done in Fig. \ref{basenetarch}, a multi-layer network of SC coding layers can be built, which minimizes the PC objective function in (\ref{predictivecodingg}). In Section \ref{clhebbian}, this paper will provide an overview of recent works studying Continual Learning systems using both SC and PC neural network architectures.

\subsection{Spiking Neural Networks and STDP Learning}
\label{snnstdp}

\begin{figure}[htbp]
\centering
    \includegraphics[scale = 0.5]{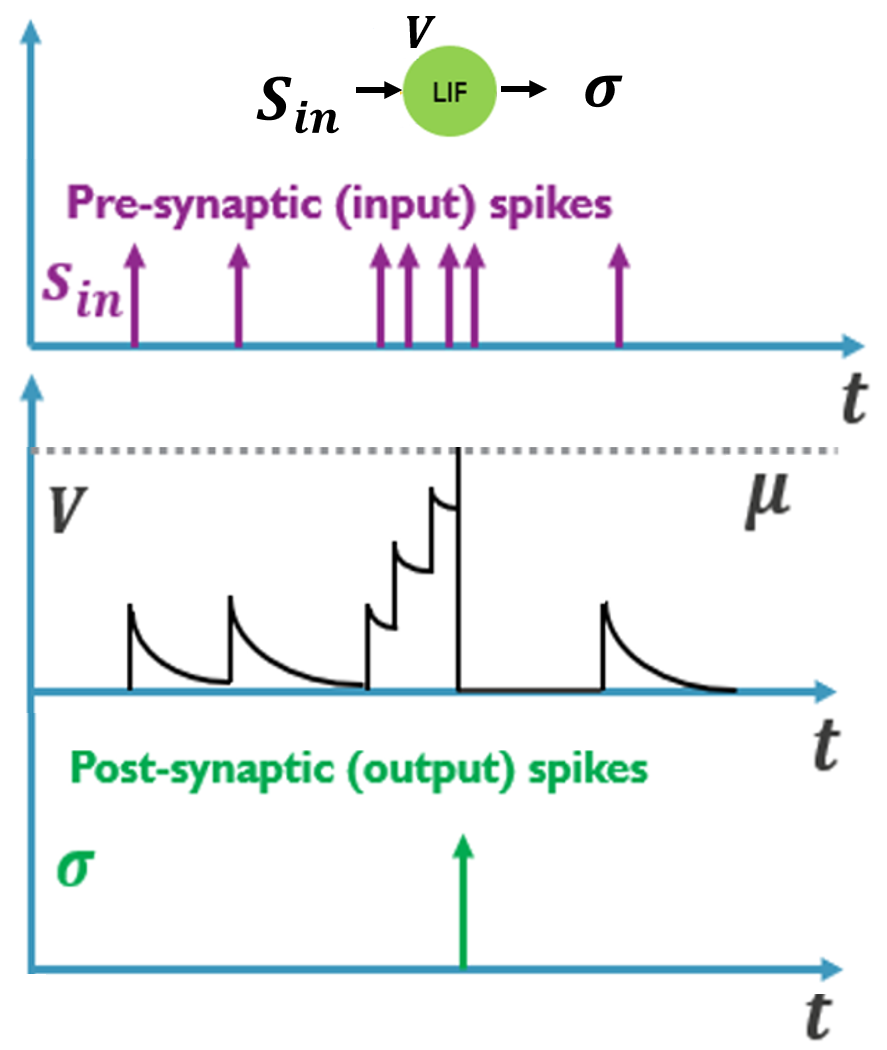}
    \caption{Conceptual illustration of the operation of the LIF neuron. In this siplified example, the LIF neuron is connected to a single spiking source $s_{in}$ via a unit weight.}
    \label{lifconcept}
\end{figure}

In contrast to the continuous activation functions used in both DNNs and the SC and PC systems presented in Section \ref{scpc}, SNNs make use of bio-plausible \textit{spiking} neurons as non-linearity, often modelled by a Leaky Integrate-and-Fire (LIF) activation \cite{10242251} (see Fig. \ref{lifconcept}):
\begin{equation}
 \begin{cases}
    \frac{dV}{dt} = \frac{1}{\tau_m} (J_{in} - V)
    \\
    \sigma = 1 \hspace{3pt} \text{\textbf{if}} \hspace{3pt} V \geq \mu \hspace{3pt} \text{\textbf{else}} \hspace{3pt} 0
    \\
    V(t+dt) = 0 \hspace{3pt} \text{\textbf{if}} \hspace{3pt} V \geq \mu
  \end{cases}
  \label{liff}
\end{equation}
with $J_{in}$ the input current to the neuron (\ref{inee}), $\sigma$ the spiking output, $V$ the membrane potential, $\tau_m$ the time constant governing the membrane potential decay, $dt$ the simulation time step, and $\mu$ the neuron threshold \cite{bookelia}. The scalar input current $J_{in}$ is continuously integrated in $V$ following (\ref{liff}). When $V$ crosses the firing threshold $\mu$, the membrane potential is reset back to zero and an output spike is emitted. The input current $J_{in}$ is obtained by filtering the inner product of the neural weights and the spiking inputs through a \textit{post-synaptic current} (PSC) kernel \cite{bookelia} (estimating the \textit{instantaneous} spiking rate):
\begin{equation}
    J_{in} = \mathcal{PSC}\{ \Bar{\phi}^T \Bar{s}_{in}(t)  \}
    \label{inee}
\end{equation}
with $\Bar{s}_{in}(t)$ the input spiking vector (originating from e.g. an event camera or other spiking neurons), $\Bar{\phi}$ the weight vector and:
\begin{equation}
    \mathcal{P}\mathcal{S}\mathcal{C} \{ x(t) \} = x(t) * \frac{1}{\tau_s} e^{-t/\tau_s}
    \label{timeconstanteffectesp}
\end{equation}
denoting the effect of PSC filtering with time constant $\tau_s$. Fig. \ref{lifconcept} conceptually illustrates the LIF neuron behavior where the single LIF neuron is connected to a pre-synaptic spiking input $s_{in}$. 

Using the LIF neuron model described above, dynamical neural networks working along the time domain can be designed using a similar connectionist approach as in conventional DNNs and their weights can be learned in order to perform useful tasks. 

\begin{figure}[htbp]
\centering
    \includegraphics[scale = 0.45]{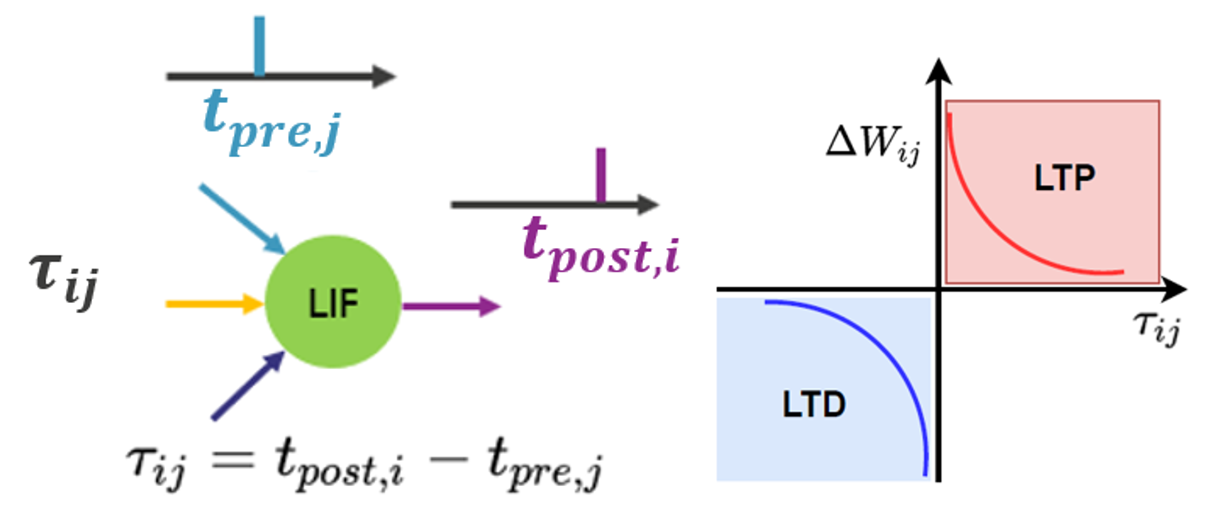}
    \caption{Conceptual illustration of STDP learning. The weight $W_{ij}$ is modified according to the double exponential STDP rule (\ref{stdpppin}) with Long Term Depression (LTD) and Potentiation (LTP) regions. 
    }
    \label{stdpconcept}
\end{figure}

Often used alongside SNNs, the study of Spike-Timing-Dependent Plasticity (STDP) is actively being pursued by the neuromorphic community as a biologically plausible and spiking local \textit{Hebbian} learning rule \cite{khacefstdp}, avoiding the high memory-access-expensive problem of backprop \cite{10.3389/fnins.2021.629892}. STDP \textit{locally} modifies the weights of each neuron following the difference $\tau_{ij}$ between the post- and pre-synaptic spike times \cite{Bi10464}: 
\begin{equation}
   W_{ij} \xleftarrow{} \begin{cases}
  W_{ij} + A_{+} e^{-\tau_{ij}/ \tau_+}, & \text{if } \tau_{ij} \geq 0
\\
   W_{ij} -A_{-} e^{\tau_{ij}/ \tau_-}, & \text{if } \tau_{ij} < 0
\end{cases}
\label{stdpppin}
\end{equation}
with $A_+, A_-$ the potentiation and depression weights, $\tau_+, \tau_-$ the potentiation and depression decay constants, $W_{ij}$ the $j^{th}$ element of the $i^{th}$ neuron weight vector $\bar{W}$, and $\tau_{ij}$ the time difference between the post- and the pre-synaptic spike times across the $j^{th}$ synapse of neuron $i$ (see Fig. \ref{stdpconcept}) \cite{Bi10464}. 

We further note as $\kappa(\tau)$ the effect of the double exponential STDP kernel on the weight modification in (\ref{stdpppin}):
\begin{equation}
    \kappa(\tau) = \begin{cases}
  A_{+} e^{-\tau/ \tau_+}, & \text{if } \tau \geq 0
\\
   -A_{-} e^{\tau/ \tau_-}, & \text{if } \tau < 0
\end{cases}
\label{stdp}
\end{equation}
The STDP rule (\ref{stdpppin}) is closely related to the Hebbian learning rule (\ref{hebbianrule}) covered in Section \ref{scpc}. Indeed, given a pre- and a post-synaptic spike train $s_{pre}(t)$, $s_{post}(t)$, the weight modification of the synapse at the post-synaptic time instant $t$ can be written as \cite{10.1371/journal.pcbi.1007835, 9892362}:
\begin{equation}
    \Delta w|_t = \eta_2 \int_{-\infty}^{\infty} s_{post}(t) s_{pre}(t - \tau) \kappa(\tau) d\tau
    \label{baseinteg}
\end{equation}
Under this form, (\ref{baseinteg}) is not useful since the post- and pre-synaptic spike trains are not known beforehand. 
Taking the expected value of (\ref{baseinteg}) over $t$ leads to:
\begin{equation}
    \mathbb{E} \{ \Delta w|_t \} = \eta_2 \int_{-\infty}^{\infty} \mathbb{E} \{s_{post}(t) s_{pre}(t - \tau) \} \kappa(\tau) d\tau
    \label{expected}
\end{equation}
The right-hand side of (\ref{expected}) can be re-written as:
\begin{equation}
    \mathbb{E}\{ \Delta w|_t \} = \eta_2 \int_{-\infty}^{\infty} \{ r_{post} r_{pre} + \mathcal{C}(\tau) \}  \kappa(\tau) d\tau 
    \label{expected2}
\end{equation}
where $\mathcal{C}(\tau)$ is the covariance between the pre- and post-synaptic spike train and $r_{post}, r_{pre}$ are the average post- and pre-synaptic spiking rates. We further rearrange (\ref{expected2}) into a rate-only term and a covariance term:
\begin{equation}
    \Delta w_{\text{STDP}} \equiv \mathbb{E}\{ \Delta w|_t \} = \eta_2 (A_+ \tau_+ - A_- \tau_-)r_{post} r_{pre} + \eta_2 \int_{-\infty}^{\infty}  \mathcal{C}(\tau)  \kappa(\tau) d\tau  
    \label{expected3}
\end{equation}

Furthermore, it can be shown \cite{sdparsecode} that the covariance term in (\ref{expected3}) can be ignored since output spikes are caused by a large number of different input spikes, making the output spike \textit{uncorrelated} to each individual input spikes due to the central limit theorem. This effect has also been experimentally assessed and confirmed in \cite{sdparsecode}.

 Therefore, the full STDP mechanism $\Delta w_{\text{STDP}}$ is approximated as follows in the remainder of the \textit{derivations} (even though full STDP is indeed used in the \textit{actual implementation} of the various SNN-STDP networks covered in this survey): 
\begin{equation}
    \Delta w_{\text{STDP}} \{ s_{post}, s_{pre}\} \approx \eta_2 (A_+ \tau_+ - A_- \tau_-)r_{post} r_{pre}
    \label{approxstdp}
\end{equation}
where it is assumed (without loss of generality) that:
\begin{equation}
    A_+ \tau_+ - A_- \tau_- = 1
    \label{conts1}
\end{equation}
in order to not modify the resulting learning speed (setting $A_+ \tau_+ - A_- \tau_->0$ is sufficient for the learning effect to take place). 
\begin{figure}[htbp]
\centering
      \includegraphics[scale = 0.55]{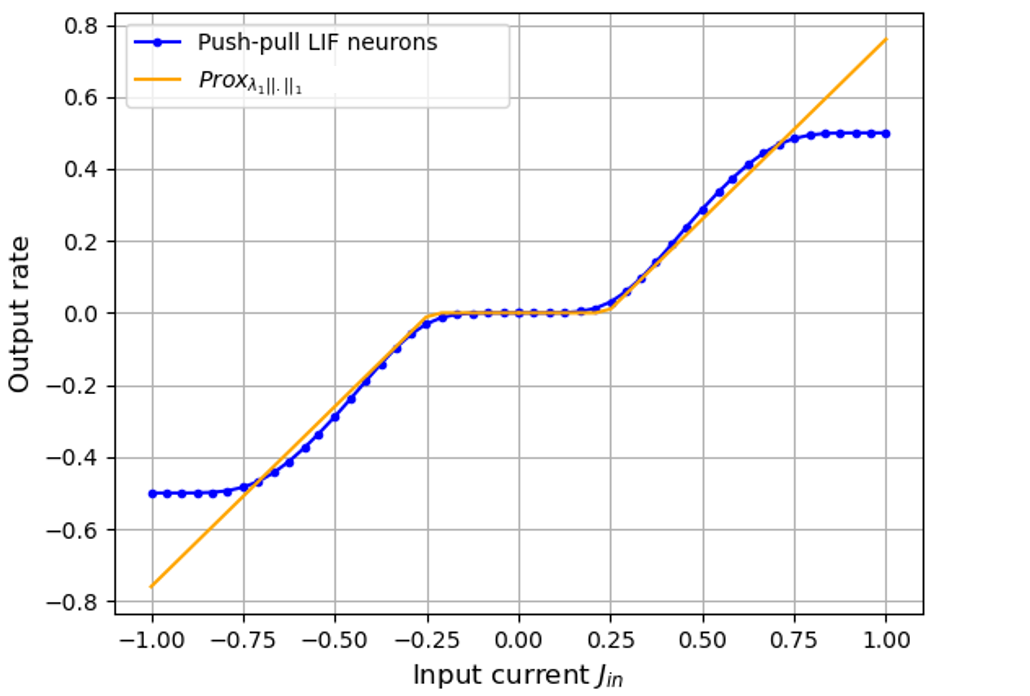}
      \caption{\textit{\textbf{Input-output relationship of a push-pull LIF neuron pair in the average spike rate domain}. Simulated using a single spiking input connected with a weight of 1 with illustration parameters $\tau_s=0.01$s, $\mu=0.15$, $\tau_m = 0.002$s in (\ref{liff}), (\ref{timeconstanteffectesp}).}}
      \label{lifconcept2}
\end{figure}

Therefore, since the effect of STDP is analogous to the effect of Hebbian learning in the local spiking \textit{rate} domain (i.e., when the local spike rate is encoding the neural activation value), STDP can be used \textit{in place} of Hebbian plasticity in the neural architecture of Fig. \ref{basenetarch}. In addition, the non-spiking neurons found in the architecture of Fig. \ref{basenetarch} can be replaced by pairs of push-pull \textit{spiking} LIF neurons (see Fig. \ref{lifconcept2}) in order to replicate the effect of the $\textbf{Prox}_{\lambda ||.||_1}$ neural non-linearity (\ref{proxop}) in the local spiking rate domain. Indeed, it was remarked in \cite{9892362, sdparsecode} that wiring two LIF activations in a push-pull setting approximates the $\textbf{Prox}_{\lambda ||.||_1}$ behavior (\ref{proxop}) in terms of the average \textit{output} spike rate in function of the average \textit{input} spike rate, with output saturation for large input rates (i.e., for the spiking probability of a time bin approaching 1) \cite{bookelia}. Fig. \ref{lifconcept2} shows the simulation result of a pair of push-pull LIF neurons compared to the shape of $\textbf{Prox}_{\lambda ||.||_1}$. 

In sum, following the observations provided in this Section about the commonalities between both STDP and Hebbian learning, and between spiking LIF neurons and the $\textbf{Prox}_{\lambda ||.||_1}$ non-linearity used in SC and PC (see Section \ref{scpc}), a number of work \cite{ororbia5, asafaslam1, asafaslam2, asafapeople} have used SNNs equipped with STDP (SNN-STDP) in a \textit{Continual Learning} setting wired following the architecture in Fig. \ref{basenetarch} (where all neurons are replaced with LIFs and Hebbian learning is replaced with STDP). This enables the extension of SC and PC to a more bio-plausible setting, using an SNN-STDP network instead of the non-spiking architecture in Fig. \ref{basenetarch}. An overview of recent works in Continual Learning with SNN-STDP will be provided in Section \ref{clstdp}.

For the sake of completeness, it is important to mention that other flavors of the STDP learning rule exists, such as reward-modulated STDP \cite{rewardstdp} or anti-Hebbian STDP \cite{asafapeople}. We refer the interested reader to the survey proposed in \cite{khacefstdp} which gives an overview of local plasticity rules in neuromorphic systems.


\subsection{The Continual Learning Problem}
\label{clproblem}

\begin{figure}[htbp]
\centering
    \includegraphics[scale = 0.9]{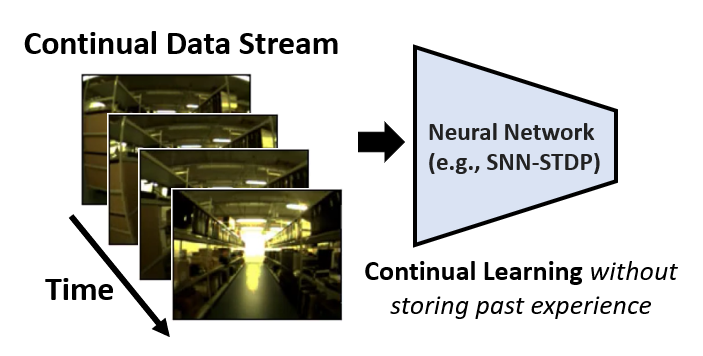}
    \caption{The continual learning scenario where a neural network (e.g., using an SNN-STDP architecture) continuously refine its weights using data points received in their natural, real-time order (without buffering and shuffling past observations in memory). At the same time, the neural network is performing inference at the edge.}
    \label{clexample}
\end{figure}

In contrast to the conventional \textit{offline} training of neural networks with training sets captured beforehand and \textit{shuffled} before being fed to the networks, continual learning is concerned with the more challenging scenario whereby a neural network must learn while being fed with a continuous stream of data presented in its natural, real-time order (see Fig. \ref{clexample}).

To illustrate this, let us consider a toy example where a neural network must learn to distinguish between class $A$ and class $B$. To achieve this, the typical method for training the network would consist of first \textit{shuffling} the training set containing both classes and then using this \textit{shuffled} dataset for training the network using a mini-batch stochastic gradient descent (SGD) approach. In contrast, let us now consider the continual learning case, where the network must learn while receiving the class $A$ and class $B$ data in a \textit{non-shuffled} order, for example, by feeding all the class $A$ data first, followed by all the class $B$ data during training. It is then clear that learning does not take place efficiently as the network will keep on over-fitting, alternating on class $A$ when all the class $A$ data are shown, followed by an over-fitting on class $B$ when all the class $B$ data are shown, effectively \textit{forgetting} what was learned using the class $A$ images. This phenomenon is famously referred to as the \textit{catastrophic forgetting} problem and constitutes a major challenge for continual learning systems \cite{9349197}.

The non-shuffled data streams found in continual-learning applications are often referred to as \textit{non-stationary} data streams \cite{nonstatdatastream} because it cannot be assumed that the prior distribution $\Gamma$ from which data points are sampled during learning stays the same throughout the learning process. Indeed, when a dataset is shuffled, data points end up being sampled in an independent and identically distributed (i.i.d) manner, as opposed to continuous streams, where subsequent data points $\Bar{o}_k$ and $\Bar{o}_{k+1}$ are strongly correlated through time $k$. In the i.i.d case of a shuffled data source, the \textit{expected value} $\mathbb{E}$ of the stochastic gradients used by SGD during DNN learning correctly converges to the gradient of the \textit{full} loss function $\mathcal{L} = \frac{1}{N_d}\sum_{k=1}^{N_d} l(\bar{o}_k)$ computed on all the $N_d$ data points:
\begin{equation}
        \mathbb{E}[\frac{\partial l(\bar{o}_k)}{\partial W}] \to \frac{\partial \frac{1}{N_d}\sum_{k=1}^{N_d} l(\bar{o}_k)}{\partial W}
    \label{reasonwhyclintro}
\end{equation}
where $W$ denotes the network weights and $l$ denotes the energy or loss to be minimized (e.g., mean square error, cross-entropy,...).

On the other hand, (\ref{reasonwhyclintro}) does not hold anymore in the continual-learning case, where the data points are sampled from a non-stationary source (i.e., a prior distribution $\Gamma$ that changes during the learning process). Since (\ref{reasonwhyclintro}) does not hold, learning is conducted \textit{outside} the assumptions made when using SGD-style techniques and hence learning becomes jeopardized. This discussion does not only hold for SGD, but also for bio-plausible local \textit{greedy descent} learning techniques like Hebbian (\ref{hebbianrule}) and STDP learning (\ref{stdpppin}), which also assume data source stationarity (see e.g., \cite{10.1162/NECO_a_00934} for a description of STDP under the light of SGD).

\subsubsection{Paradigms and methods for Continual Learning}
\hfill \break
\label{paradigmscl}

Continual Learning tasks can broadly be separated in two different categories. The first category which is commonly referred to as \textit{task-incremental} continual learning denotes the scenario where a neural network must learn $N_T$ different tasks $\mathcal{T}_{1,...,T}$ (e.g., $\mathcal{T}_1$: discriminating between cat and dog images,  $\mathcal{T}_2$: discriminating between plane and car images, etc.). The learning of these $T$ different tasks is done in an incremental manner, where the neural network is first trained on $\mathcal{T}_1$, then trained on $\mathcal{T}_2$, and so on. Hence, the goal during learning is to be able to learn new tasks while \textit{not forgetting} the old tasks that were already learned. It is important to note that during the learning of each task \textit{in isolation}, the neural network could have access to a shuffled i.i.d set of data for this particular task \cite{9349197}. In this case, the non-i.i.d nature of the learning comes from the fact that the data distribution \textit{across} the tasks will exhibit a large variability (e.g., $\mathcal{T}_1$ containing shuffled images of cats and dogs, and $\mathcal{T}_2$ containing shuffled images of planes and cars).  

Under this \textit{task-incremental} scenario, the boundaries between the different tasks are known. This is in contrast to the second broad category of continual learning paradigm which is referred to as \textit{task-free} continual learning. In the \textit{task-free} scenario, the boundaries between tasks, categories or data sources is unknown during the continual learning process. In this second, more challenging scenario, the continual learning system must learn using the real-time data stream which is usually heavily unbalanced and non-i.i.d. For instance, a notable line of work in the task-free setting consists of having robot agents learning to encode visual sensory data as they explore an unknown environment, for performing tasks such as Simultaneous Localization and Mapping (SLAM) \cite{clslam, asafaslam1, asafaslam2}.

Regarding methods for alleviating the catastrophic forgetting issues encountered during continual learning, most research has focused on three broad categories of approaches, usually studied in the \textit{less bio-plausible} DNN context. The first category is referred to as \textit{replay methods} and consists of keeping a limited amount of data from each task seen in the past, or by keeping segments of past sensory streams in order to feed them again for training in the future. Hence, this method seeks to alleviate catastrophic forgetting by episodically replaying past training data. Even though useful, this method requires potentially significant memory overheads, with a growing memory size to store all past experiences. This makes it less suited for continual learning in resource-constrained neuromorphic edge devices where memory consumption is usually limited. A second category of method is referred to as \textit{parameter isolation methods} \cite{9349197} and seeks to dynamically add new architectural elements to the neural network each time a new task is encountered. Hence, this method is more suited for task-incremental continual learning scenarios. In addition, this method also suffers from dynamically growing memory consumption, making it also less suited for energy- and area-efficient neuromorphic computing in edge devices. Finally, the third broad category of continual learning methods found in DNN research is commonly referred to as \textit{regularization methods}. Regularization methods use weight regularization techniques that first identify the \textit{importance of each weight} for the task that have been learned up to the present point in time. Then, the more a weight is considered important, the less the weight regularization enables that weight to change during the learning of new future tasks. In contrast to the previously mentioned method, the regularization method does not lead to significant memory overheads, and has been demonstrated in both task-incremental and task-free settings \cite{taskfreeMAS}.

More closely related to ANNs with Hebbian learning and SNNs equipped with STDP, a number of prior work \cite{aljundi2018selfless} have shown that providing \textit{activation sparsity} in DNNs using lateral inhibition between the neurons of each layer (as in the bio-plausible SC and PC systems covered in Sections \ref{scpc} and \ref{snnstdp}) has a beneficial effect on continual-learning performance, usually leading to a significant boost in the network accuracy after learning. Related to this observation, it has been remarked that gradient sparsity during surrogate gradient-based learning can also have a beneficial effect on continual learning in SNNs \cite{elephant}. In addition, other works have shown the links between popular regularization techniques used for CL in Deep Learning contexts, with Hebbian learning theory \cite{memoryaware}. 

Hence, the study of bio-plausible neural networks built following the SC and PC approaches using local learning rules such as Hebbian and STDP learning has attracted a growing attention for the design of continual-learning systems for extreme edge applications. To this end, the next Sections of this paper will provide an overview of the recent research done in the design of bio-plausible continual learning systems.


\section{Continual learning in Hebbian networks}
\label{clhebbian}
\subsection{Task-incremental Continual Learning}
\label{taskincrnonspike}
In a recent pioneering work, A. Ororbia \textit{et al.} \cite{ororbia1} have proposed a lifelong continual learning system based on Predictive Coding (see Section \ref{predictivecode}), and have demonstrated their findings on common \textit{task-incremental} classification problems (see Section \ref{paradigmscl} for a description of the task-incremental setting). It is important to note that even though demonstrated on task-incremental benchmarks, the network proposed in \cite{ororbia1} can work in a task-free setting, as it has the intrinsic capability of detecting if an incoming sample originates from a new task without receiving external task identifiers. This forms a particularly innovative aspect of this work.

The network proposed in \cite{ororbia1} is built as follows using two distinct neural ensembles. The first neural ensemble is a Predictive Coding network as described in Section \ref{predictivecode}. This Predictive Coding network acts as a generative model which can reconstruct its input from its output sparse code (i.e., by re-projection following $o\approx \Phi c$ in (\ref{predictivecodingg})). The Predictive Coding network is fed as input a vector $s_{in} = \{o, y, t\}$, combining the input signal $o$, its one-hot class vector $y$ and a \textit{task-pointer} vector $t$. The task pointer vector $t$ is produced by a second neural ensemble and helps the predictive coding backbone at better selecting different sub-networks routes for each different task. After the continual learning phase has ended, classification of test data can be done by feeding the input observation only and reconstructing (by re-projection $o\approx \Phi c$) the estimated label $\hat{y}$.

The task-pointer vector $t$ is produced by a second neural ensemble referred to as the \textit{task selector} network. The motivation behind this design choice comes from neuroscience where it is believed that the basal ganglia plays a role in information routing in the brain, selecting different programs and memory stored in other regions in the cortex. The task selector network detect the beginning of new tasks by monitoring the running re-projection error of the Predictive Coding network in (\ref{predictivecodingg}). If this running re-projection error exceeds its previous value by at least twice its running standard deviation, a task boundary is detected. This information is used by the task selector to infer a task-pointer vector $t$ from the observation $o$ using a SC Hebbian network. 

The authors in \cite{ororbia1} benchmark their proposed system against a large number of conventional continual learning methods used in the Deep Learning field, such as Elastic Weight Consolidation (EWC) \cite{EWC}, Synaptic Intelligence (SI) \cite{SI}, Memory Aware Synapses (MAS) \cite{memoryaware} and so on. Experiments are conducted using three common benchmark datasets extensively used in continual learning: \textit{Split MNIST, Split NotMNIST} and \textit{Split Fashion MNIST} \cite{ororbia1}. These datasets are based on the orginal MNIST, NotMNIST and Fashion MNIST, but are shown in a class-incremental manner, each data class at a time (without revisiting past classes). It is shown in \cite{ororbia1} that their proposed method is the top performer in terms of classification accuracy compared to the other competing continual learning methods that were assessed.


\subsection{Continual Sequence Modelling}
In conventional deep learning systems, sequence modelling is usually carried using Recurrent Neural Networks (RNNs) trained using Back-propagation through time (BPTT). Although useful in the deep learning context, BPTT is a highly compute-expensive, non-local learning rule with little bio-plausibility. This is why an emerging body of work \cite{annabi,ororbia2} are studying the use of PC networks for continual sequence modelling, without resorting to BPTT as their learning rule \cite{bptt}.

To help uncover roads towards the bio-plausible continual learning of time sequences, L. Annabi \textit{et al.} \cite{annabi} have proposed an innovative continual sequence learning system based on Predictive Coding. In particular, the authors in \cite{annabi} take inspiration from Predictive Coding to build an RNN and study a wide range of different methods for learning the input weights, the recurrent weights and the output weights of their proposed 1-hidden-layer RNN architecture. Their system embarks the predictive coding mechanism for learning the recurrent and input weights of their 1-hidden-layer RNN, while experimenting with a range of non-predictive coding techniques for the output weights (e.g., using EWC \cite{EWC} for the output weights, among other techniques tried by the authors). Experimental demonstrations are provided using motion-capture trajectories of dimension $62$ from the CMU Motion Capture database, denoting the evolution of the joints of a human body in the environment. In addition, the authors also provide experimental demonstrations using the 2D handwritten letter trajectories from the UCI Machine Learning repository \cite{ucidatabase}. First, the author go on to benchmark different \textit{output weight} learning method and demonstrate that the use of the \textit{Conceptor} method proposed in \cite{conceptor} outperforms both the use of EWC and stochastic gradient descent in terms of sequence prediction error for the learning of the output weights. Then, the learning of the \textit{recurrent weights} is studied, and the authors remarkably show that incorporating learning in the recurrent weights following their predictive coding strategy only slightly improves the sequence prediction error compared to the use of a randomly-initialized RNN with random recurrent weights. Finally, the authors experiments with different \textit{input weight} learning approaches and show that the proposed Predictive Coding approach outperforms the use of stochastic gradient descent. In sum, the author conclude that their work show the usefulness of combining Predictive Coding and the Conceptor method \cite{conceptor} for output weight regularization, since this combination lead to the lowest sequence prediction error in their experiments.

Related to the work in \cite{annabi}, A. Ororbia \textit{et al.} \cite{ororbia2} have proposed a novel bio-plausible framework for continual learning in RNNs by locally aligning distributed representations captured by each layer of a standard Predictive Coding architecture, equipped with Hebbian learning. Their method, termed \textit{Parallel Temporal Neural Coding Network} (P-TNCN) is essentially based on the Predictive Coding approach (covered in Section \ref{predictivecode}) where the first layer receiving an input $o_k$ at time-step $k$ seeks to predict the input for the next time-step $o_{k+1}$ while producing a hidden state $c^1_k$. Then, the next layer receiving $c^1_k$ seeks to predict the next value of the first hidden state $c^1_{k+1}$ by producing a second hidden state $c^2_k$, and so on for as much layers forming the network. In addition to the Hebbian learning rules already found in the Predictive Coding approach, the authors in \cite{ororbia2} augment the Predictive Coding Hebbian mechanisms with an additional Hebbian term (\ref{addhebb}) which seeks to improve the generalization capability of the network by promoting the learning of a model which also exhibit some level of co-occurrence between the statistics of each layer, while minimizing the predictive coding error as well.  
\begin{equation}
    \Delta_{\text{HebbReg}}W = -\frac{c_k^l (c_{k-1}^{l-1})^T}{||c_k^l (c_{k-1}^{l-1})^T||_2}
    \label{addhebb}
\end{equation}
where $l$ denotes the layer index and $k$ the time step.

Then, the authors in \cite{ororbia2} propose an extensive experimental study to demonstrate the performance of their proposed P-TNCN method by comparing it against two popular methods used for training RNNs in an online fashion (RTRL \cite{rtrl} and UORO \cite{uoro}). In addition, the proposed system is also compared against the use of \textit{Echo State Networks} (ESN) which are a class of RNNs with untrained recurrent weights and trained output weights. Four different datasets are used during the experiments: \textit{i) Bouncing MNIST, ii) Bouncing NotMNIST, iii) Bouncing Fashion MNIST and iv) Penn Treebank}. In the Bouncing MNIST, each data point is a 20 frames long sequence containing two digits from the standard MNIST dataset randomly bouncing around a $64 \times 64$ patch. The Bouncing NotMNIST and Fashion MNIST follow a similar construction, while the Penn Treebank dataset contains sequences of words (sentences) for studying Natural Language Processing (NLP). 

Remarkably, the authors in \cite{ororbia2} show that their proposed Hebbian-learning P-TNCN system outperforms a number of competing methods by reaching a lower loss on the test set for all bouncing datasets. Their proposed algorithm only falls short compared to BPTT when testing on the Penn Treebank dataset, possibly due to the fact that the text data is highly discontinuous. In addition, the authors show the applicability of their proposed methods for zero-shot adaptation to out-of-sample data, and for the online continual learning of data sequences via generative modelling. In sum, to the best of our knowledge, the work proposed by A. Ororbia \textit{et al.} \cite{ororbia2} constitutes one of the most advanced account on the use of ANNs equipped with Hebbian learning for the challenging task of continual sequence modelling.



\subsection{Dynamical agent control via Hebbian networks}

An emerging number of work have shown the applicability of Hebbian networks for agent control tasks that have been previously solved with Reinforcement Learning techniques. A general trend among these works is the fact that learning can be done with either extremely sparse rewards, or either no reward mechanism at all. This is in contrast to traditional methods used in Deep Reinforcement Learning which usually rely on the availability of dense rewards from the environment. The reason behind this phenomenon lies in the fact that these recently-proposed Hebbian agent control systems are built by taking inspiration from the Active Inference framework \cite{activeinferencebook}, which achieves agent control by learning in an \textit{unsupervised} manner a \textit{generative model} of the agent's environment $P(o,a,s)$, where $o$ is the agent's observation (e.g., position, camera view, etc.), $a$ is the action performed by the agent and $s$ is the latent state of the agent (i.e., a vector summarizing the state of the agent in the environment). Hence, by learning in an unsupervised manner a probability distribution $P(o,a,s)$ over observations, actions and states, an agent can steer itself towards its goal by determining which sequence of actions $\Tilde{a} = \{a_1,...,a_L \}$ will maximize the probability $P$ of reaching the goal state $s^*$, with associated observation $o^*$ \cite{activeinferencebook}.

Using Sparse Coding, A. Safa \textit{et al.} \cite{safaIWAI} have proposed a continual-learning Hebbian Active Inference system and demonstrated its application using the Mountain Car environment from the Open AI gym suite. Crucially, the authors have shown that their proposed system reaches two orders of magnitude faster convergence rate compared to Q-learning (typically used to solve this task), while not requiring a replay buffer as in traditional reinforcement learning systems. Their proposed Hebbian system architecture is built as follows: a first sparse coding network is used to encode the observation data $o$ (car's position and velocity) and action vector $a$ (car's acceleration) into a latent space $s$, corresponding to the output sparse code of the Hebbian network in Fig. \ref{basenetarch}. Then, a second Hebbian network is set up following an auto-regressive approach, which estimates the next future state $s_{k+1}$ given a number of past states $\{s_{k-K},...,s_k\}$ and the corresponding actions that were performed by the agent $\{a_{k-K},...,a_{k}\}$. By learning this model, the agent can perform \textit{planning} in order to steer itself towards its desired goal (the top of the mountain in the case of the Mountain Car environment). This planning is done by randomly sampling sequences of actions $\Tilde{a} = \{a_{1},...,a_{L}\}$ and estimating the resulting sequence of states $\Tilde{s} = \{s_{1},...,s_{L}\}$ using the auto-regressive Hebbian network. Finally, the optimal action sequence $\Tilde{a}^*$ that minimizes the distance between the state sequence $\Tilde{s}$ and the goal state $s^*$ to be attained is selected and executed by the agent.

In a series of pioneering work using Predictive Coding, A. Ororbia \textit{et al.} \cite{ororbia3, ororbia4} have proposed a bio-inspired robot control framework termed Active Predictive Coding (ActPC) and have successfully applied the proposed technique on a range of different robot control tasks from both the \textit{Mujoco} control suite (\textit{Reacher-v4, Half-Cheetah-v2, Hopper-v2, Walker-v2 and Swimmer-v2}), and from the challenging \textit{Robosuite} environment (\textit{Block Lift} and \textit{Can Place} tasks). By experimentally comparing their proposed system to a range of conventional reinforcement learning techniques, the authors show that their proposed ActPC system is either on-par or either outperforms the use of conventional reinforcement learning techniques while being bio-plausible and only resorting to local Hebbian learning rules. Different to the Hebbian Active Inference system described earlier \cite{safaIWAI}, their system architecture follows an \textit{actor-critic} approach where a first Predictive Coding network plays the role of \textit{actor}, proposing the next action given the current observation. Then, a second Predictive Coding network plays the role of \textit{critic}, estimating the expected reward based on the observation and the action proposed by the \textit{actor} network. These two networks are complemented with additional Hebbian learning mechanisms for encouraging environment exploration and generating goal-seeking behavior, in order to balance the exploration-exploitation trade off (i.e., how much the agent is learning new behavior and how much the agent is seeking to reach the goal). 

\subsection{Continual Simultaneous Localization and Mapping via Hebbian learning}

Simultaneous Localization and Mapping (SLAM) is a fundamental task for robot navigation, where a robot must jointly construct a map of the environment in which it is navigating while localizing itself into that map \cite{ratslam}. Learning-based SLAM systems are usually built by first training a neural network model to perform the visual encoding of sensory data (e.g., RGB camera) into feature descriptors. Then, these feature descriptors are used to detect new robot poses and locations already visited by the robot in order to perform \textit{loop closure} detection. Once a loop closure is detected, the map that is being built by the SLAM algorithm can be refined and the robot position corrected. But these learning-based SLAM approaches rely on pre-trained neural network using training sets acquired beforehand. This comes in contrast with the fact that these SLAM systems must work in new environments, often \textit{not} captured by static training set acquired \textit{a priori}.

For this reason, there has recently been a growing interest in using continual learning with SLAM, for learning the neural network encoder on the fly, as the robot is exploring unseen environments \cite{clslam}. Following this line of research, A. Safa \textit{et al.} \cite{asafaslam1} have proposed a continual learning SLAM system using a Sparse Coding-based Hebbian learning network. The key innovation of their proposed system lies in the use of a Hebbian network similar to the network shown in Fig. \ref{basenetarch} in order to encode the input RGB image into lower-dimensional feature descriptors (of dimension $64$). Then, the authors in \cite{asafaslam1} feed the obtained descriptor into a RatSLAM back-end \cite{ratslam} which gradually builds a map and localize the agent by using the feature descriptors for loop closure detection, correcting the raw odometry provided by an Inertial Measurement Unit (IMU). Crucially, the authors in \cite{asafaslam1} provide a \textit{surprise-modulated} Hebbian learning rule for reaching better continual-learning performance. Each time a new camera observation $o_k$ is received, the following \textit{surprise factor} is computed:
\begin{equation}
    \mathcal{S} = || \Phi_{k-1} c_k - o_k ||_2^2 - || \Phi_{k-1} c_{k-1} - o_{k-1} ||_2^2
    \label{suprise}
\end{equation}

If the surprise factor in (\ref{suprise}) is positive ($\mathcal{S} > 0$), the input observation $o_k$ at the current time step $k$ is used to perform a Hebbian learning step (Eq. \ref{hebbianrule1} and \ref{hebbianrule2}), refining the dictionary (or weight matrix) $\Phi$. Else, no learning is executed for that time step.

Intuitively, this technique enables the filtration of incoming data points that are deemed redundant and already learned by the system. Indeed, if the incoming input observation $o_k$ leads to a re-projection error $|| \Phi_{k-1} c_k - o_k ||_2^2$ that is \textit{larger} than the previous re-projection error $|| \Phi_{k-1} c_{k-1} - o_{k-1} ||_2^2$ at time step $k-1$, this indicate that the incoming observation has not yet been well captured by the network weights $\Phi$ and hence must be incorporated into the learning process. On the other hand, if $\mathcal{S} < 0$, this means that the incoming observation is probably already well captured by the Hebbian network. Hence, integrating that observation point may lead to the over-fitting of the network on the local data stream and lead to catastrophic forgetting (see Section \ref{clproblem}). 

The authors in \cite{asafaslam1} experimentally validate their findings using a drone flying inside a warehouse environment, by showing that the proposed surprise-driven Hebbian learning method (\ref{suprise}) leads to a lower replay error compared to the use of standard Hebbian learning. Furthermore, the authors also show that the proposed surprise-driven method achieves better network stability compared to the use of standard Hebbian learning, where it was observed that the learning process could diverge in some particular network settings.

Finally, the authors use the proposed continual Hebbian learning setup to demonstrate a complete SLAM system using a drone flying between storage racks and open areas. Experimental results show that the proposed system reaches either on-par or either better performance compared to state-of-the-art SLAM systems, while not requiring any offline network pre-training nor any explicit training data acquisition step. Hence, the work in \cite{asafaslam1} opens the door to the application of Continual Hebbian learning for performing SLAM in unknown environment not captured by datasets available beforehand.

\begin{table}
\caption{\label{jlab1}Works in the field of continual Hebbian learning covered in Section \ref{clhebbian}.}
\footnotesize
\centering
\begin{tabular}{@{}lll}
\br
Reference & CL Application & Architecture \\
\mr
A. Ororbia \cite{ororbia1} & Task- and Class-Incremental &Predictive Coding \\

L. Annabi \textit{et al.} \cite{annabi} & Sequence Modelling & Predictive Coding \\

A. Ororbia \textit{et al.} \cite{ororbia2} & Sequence Modelling & Predictive Coding \\

A. Safa \textit{et al.} \cite{safaIWAI} & Agent Control &Sparse Coding \\

A. Ororbia \textit{et al.} \cite{ororbia3} & Agent Control &Predictive Coding\\

A. Ororbia \textit{et al.} \cite{ororbia4} 
 & Agent Control & Predictive Coding \\
 
A. Safa \textit{et al.} \cite{asafaslam1} & SLAM &Sparse Coding\\

\br
\end{tabular}\\

\end{table}

Table \ref{jlab1} briefly summarizes the works in the field of continual Hebbian learning that have been covered in this Section. In the next Section, an overview of continual-learning works using SNNs with STDP learning will be presented.

\section{Continual learning in SNNs with STDP}
\label{clstdp}

\subsection{Task-incremental Continual Learning with SNNs and STDP}

In a recent paper, D. I. Antonov \textit{et al.} \cite{antonov} have proposed a first study on how to achieve continual learning with SNNs trained with local STDP rules. A key aspect of their work is that they develop a continual-learning method for determining the importance of each weight in their network using \textit{stochastic Langevin dynamics}, in contrast to the use of gradient-based regularization techniques such as EWC, SI and so on \cite{EWC, SI}, which lack bio-plausibility due to their non-local nature. The network used in their experiments uses two local learning rules: conventional unsupervised STDP and the \textit{supervised} reward-modulated STDP which potentiate the weights when a reward is received (e.g., if the network made the correct prediction) and vice versa in the opposite case. The network architecture consists of three convolutional layers followed by pooling layers, using Integrate and Fire (IF) spiking neurons \cite{aliradar}. The last layer is used to perform classification where the neuron with the earliest spike time denotes the inference decision.  

In order to achieve continual learning, the authors in \cite{antonov} experiment with a number of methods such as; \textit{i)} using winner-take-all (WTA) competition mechanisms between the neurons of each layer (following indications in literature about the importance of sparsity for continual learning \cite{rahaf}); \textit{ii)} using pseudo-rehearsal where noise patterns are fed to the network and the output is recorded for being used as a pseudo-rehearsal technique for when a new task must be learned; \textit{iii)} using regularization methods by applying noise during training, using dropout \cite{dropout}, freezing the large weights during learning, and using their proposed \textit{Langevin dynamics} technique. 

The proposed Langevin dynamics method for continual learning seeks to determine the domain in which each weight $w_i$ can be changed while not affecting the accuracy of the system. By determining this domain $D_i$, each weight $w_i$ can then be constrained to stay within its particular domain $D_i$ when a new task is being learned, reducing catastrophic forgetting on the previous task. To find such multi-dimensional region of permitted weight values, Langevin dynamics is introduced by applying noise to the weights. During their random exploration of the weight space, when the weights enter a region where the network accuracy is dropping, the R-STDP mechanism then acts as feedback mechanism bringing the weights values back into the admitted region. After enough time, the Brownian motion of the weights will explore the permissible region, which is used to constraint the weights during the learning of new tasks.

Then, the authors in \cite{antonov} perform a series of experiments in a task-incremental setting using the MNIST and the Extended MNIST (EMNIST) dataset (containing both letters and digits). The task-incremental learning setup is conducted as follows: first, the network is trained using MNIST and then, assuming the MNIST data is not available anymore, the network is then trained using the EMNIST data as the second task. The authors compare the various continual learning techniques explored in their work in terms of global accuracy on both tasks. It is shown that their proposed approaches using Langevin dynamics is highly promising and came close to the performance achieved using \textit{self-reminder} techniques (replay using a subset of the previous task) while being \textit{less memory intensive} than self-reminder techniques and hence, better suited for continual learning application in emerging neuromorphic edge devices. Indeed, Langevin dynmics and related Markov Chain Monte Carlo (MCMC) methods have been found as well-suited for implementation in neuromorphic edge devices though the innovative use of memristor technology, by exploiting the inherent stochasticity of these emerging hardware devices \cite{memristormcmc}.

In another work by A. Ororbia \cite{ororbia5}, the author proposes a continual learning SNN architecture based on Predictive Coding using an STDP-like Hebbian local learning rule. The framework, termed \textit{spiking neural coding}, can be seen a spiking implementation of the Predictive Coding network architecture proposed by the same authors in \cite{ororbia1} and discussed in Section \ref{taskincrnonspike} (although their SNN implementation does not feature the \textit{task selector} network used in \cite{ororbia1}). Their SNN forms a generative model which receives as input tuples $\{X, y\}$ containing the data points $X$ and associated one-hot encoded label $y$, converted to spike trains using Poisson coding. Then, at test time, only the test data $X_{test}$ is presented to the network which can then estimate the associated label $\hat{y}$ by re-projection of the neural activity into the input space ($o\approx \Phi c$ in (\ref{predictivecodingg})).

Then, extensive experimental demonstrations are provided to benchmark the proposed spiking neural coding architecture against a number of state-of-the-art deep learning techniques used for alleviating catastrophic forgetting (such as EWC, SI and so on \cite{EWC, SI}). Experiments are conducted using datasets such as MNIST, Fashion MNIST, Stanford Optical Character Recognition and Caltech 101 Silhouettes \cite{mnist, fashionmnist, caltech101}. After demonstrating the usefulness of their proposed algorithm under the conventional training procedure (i.e., not in a continual learning manner), the author in \cite{ororbia5} provide a study of the memory retention ability of their proposed SNN \textit{under the continual learning setting} this time. It is remarkably shown that their proposed spiking neural coding approach obtains highly competitive continual learning performance compared to state-of-the-art methods, outperforming them on both MNIST and Fashion MNIST datasets. 

Finally, another SNN architecture for continual learning has been proposed by J. M. Allred \textit{et al.} \cite{alfredo}, which differs from the previously covered works through its neuroscience-based use of dopaminergic plasticity modulation for achieving \textit{unsupervised} continual learning. The proposed framework, termed Controlled Forgetting Networks (CFNs), temporarly makes the synaptic weights of some neurons more plastic (i.e., more subject to change), while keeping fixed the weights of other neurons. This allows an isolated and distributed adaptation of the SNN weights through STDP. This selective modulation is achieved by taking close inspiration from biology. In the brain, dopamine acts as a neuromodulator on synaptic plasticity, where dopamine releases are associated with the encountering of novel, surprising data points. This surprise-driven synaptic modulation is achieved by scaling the STDP learning rate of each neuron through a set of additional \textit{dopaminergic} neurons that take as input the spiking activity of neurons in each layer and feed back their output to each neuron in that layer in order to modulate their individual learning rate.

Experimental results are provided using the MNIST dataset and it is shown that the proposed method successfully achieves continual learning, retaining an accuracy larger than $90\%$ after learning each digit class in MNIST one by one. In addition, the authors conduct ablation studies to show the usefulness of their various design choices and their newly-proposed dopaminergic neurons, where not using these aforementionned techniques lead to a severe decrease in accuracy during the continual learning process. 

Hence, the works discussed above show the usefulness of SNNs in continual learning settings and strongly motivates their exploration in other continual learning contexts, beyond task-incremental learning.

\subsection{Continual SNN-STDP learning for Robotics applications}

A number of SNN-based works have been studied in literature for performing Simultaneous Localization and Mapping (SLAM) \cite{snnslamqut1, snnslamqut2}, but these work usually rely on the offline training of the SNN visual encoder networks, instead of using continual learning, which is the focus of this survey.

In order to extend SNN-based robotics systems to the continual learning case, a number of recent works have been exploring the use of SNN equipped with \textit{continual} STDP learning in robotics applications, with a focus on solving various drone navigation tasks such as Simultaneous Localization and Mapping (SLAM) and people detection from drones.

A first-of-its-kind SLAM system has been proposed by A. Safa \textit{et al.} \cite{asafaslam2} which can continuously learn to encode sensory data on the fly into feature descriptors (using an event-based camera and a radar as input sensors to their SNN-STDP network). The authors in \cite{asafaslam2} set up an SNN-STDP architecture which essentially acts as a spiking substrate of computation for Sparse Coding (introduced in Section \ref{sconly}). Upon receiving the spiking sensory data, the SNN-STDP continuously refines its weights on the fly as the drone is flying in an indoor warehouse-like environment and jointly produces a spiking output sparse code. These output spike trains are averaged over a time window $\Delta T = 50$ ms to obtain continuous-valued feature descriptors. Finally, the produced feature descriptors are fed to a RatSLAM back-end \cite{ratslam} for performing loop closure detection and map correction.

Then, the authors in \cite{asafaslam2} benchmark their proposed continual SNN-STDP SLAM algorithm against a number of CNN-based and feature-based SLAM approaches by mounting an indoor Ultra Wide Band (UWB) positioning beacon on the drone for acquiring ground truth localization and mapping data. It is shown that the proposed approach either reaches on-par or either outperforms the competing SLAM methods while being extremely robust to lighting variation, due to its use of a radar sensor coupled with an event-based camera (which typically performs well in low-light due to its high dynamic range \cite{surveydvs}).

\begin{figure}[htbp]
\centering
    \includegraphics[scale = 0.4]{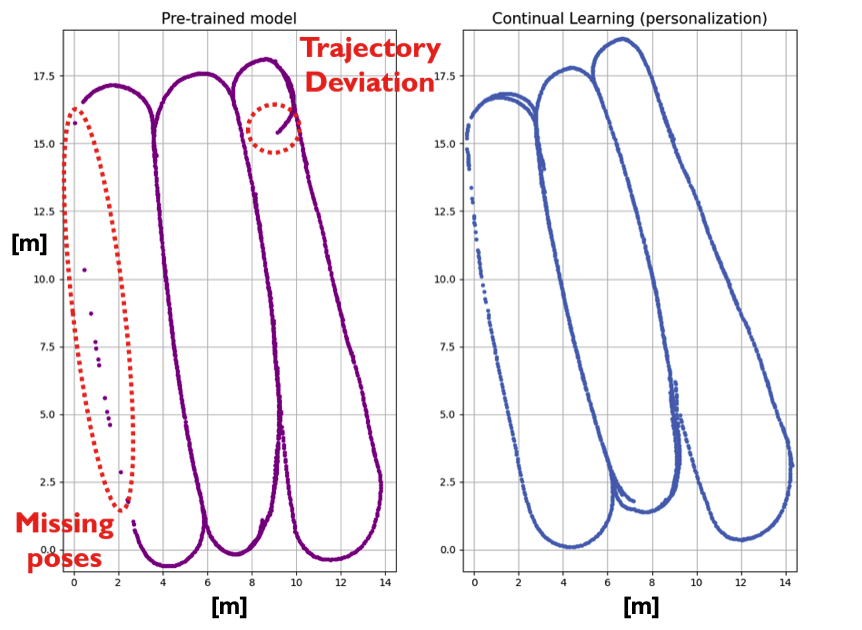}
    \caption{Pre-training vs. Continual Learning for the continual learning SNN-STDP SLAM system proposed in \cite{asafaslam2, asafaslam3}. The system trained offline suffers from a significant number of missing poses and trajectory deviations. The axis on the graph have dimensions in meters [m].}
    \label{offlinevscl}
\end{figure}

A remarkable aspect of their system is the fact that the SNN is never pre-trained on an offline acquired dataset. Rather, their SNN-STDP network is initialized randomly and continuously refines its weights on the fly as the drone explores new unknown environments. Interestingly, the authors show in \cite{asafaslam3} that the \textit{continual learning nature} of their proposed SNN-STDP system brings novel value from an algorithmic point of view, by outperforming the same SNN-STDP architecture but \textit{trained offline} using a pre-recorded dataset (see Fig. \ref{offlinevscl}).

This first demonstration of a continual learning SLAM system using SNN-STDP opens the door for the conception of novel robots that can learn to navigate in unseen environment without resorting to training sets acquired offline.

\begin{figure}[htbp]
\centering
    \includegraphics[scale = 0.6]{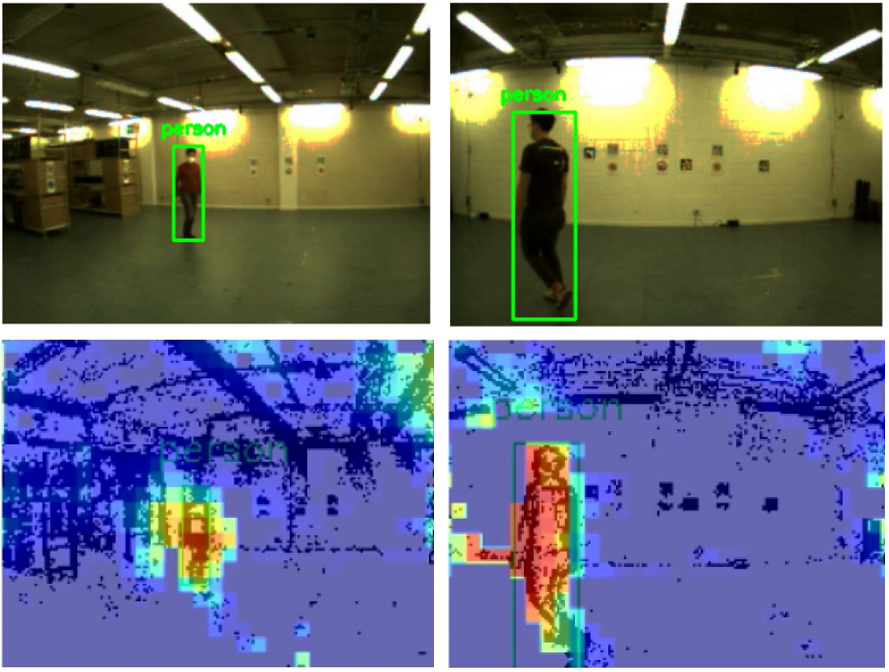}
    \caption{Attention maps from the DVS data, inferred by the continual-learning SNN-STDP system in \cite{asafapeople}. The label bounding boxes are also shown.}
    \label{floril}
\end{figure}

In another work by A. Safa \textit{et al.} \cite{asafapeople}, an attention-based people detection system has been proposed for detecting potential human subjects walking in front of a flying drone in order to provide collision avoidance. Their proposed system makes use of an innovative continual learning SNN-STDP network built following a fully-convolutional network architecture. Indeed, a fully convolutional architecture is utilized in order to provide \textit{translational invariance} (i.e., detecting people in an invariant manner regardless of their position on the image plane). The continual learning process of their system works as follows. First, a drone embarking an event-based camera (which provides spiking image data) is made to fly in an indoor environment where human subjects are walking in front of the drone. During this flight, the SNN-STDP system is continuously learning to output an attention map where high pixel values denote the presence of a human subject (see Fig. \ref{floril}). After a $\sim 2$ minute learning sequence, the SNN weights are fixed and the complete system is assessed on other flight sequences in order to measure the detection performance of the proposed people detection architecture. 

During the continual learning phase, the SNN learning is done through a semi-supervised STDP-driven setup, where a supervised STDP contribution modifies the SNN weights using the human location labels, and an unsupervised STDP contribution is used as a regularization method providing better detection performance by reducing the amount of false alarms in the produced attention maps. Remarkably, the authors in \cite{asafapeople} show that their proposed continual learning SNN-STDP system outperforms the use of a deep CNN \textit{trained offline} by $+19\%$ in terms of human detection performance using the event-based camera data. It is argued that this superior performance is linked to the recurrent nature of the proposed SNN-STDP system which better integrates the timing information embedded in the event-based camera data stream.

Together with the SLAM system proposed in \cite{asafaslam2}, the SNN-STDP-based people detection system presented in \cite{asafapeople} constitute what is, to the best of our knowledge, some of the most advanced real-world applications of continual SNN-STDP learning demonstrated in the robotics context. 

\subsection{Continual SNN-STDP implementations in Integrated Circuits}

An important direction for continual SNN-STDP systems lies in the design of novel Integrated Circuits (ICs) or chips that can perform continual learning at the extreme edge with reduced energy and area utilization. Indeed, building such novel ICs will enable the conception of a novel range of IoT and robotics devices that can be personalized on the fly to their particular environment.

Within this context, I. Munoz-Martin, S. Bianchi \textit{et al.} \cite{chip1, chip2} have studied in a number of works the conception of a continual learning system using unsupervised STDP learning implemented in hardware using CMOS logic and phase change memory (PCM) circuits for implementing the SNN synapses. The authors in \cite{chip1, chip2} build a mixed supervised-unsupervised network comprising a supervised convolutional network feeding into an unsupervised STDP network with Winner-Take-All behavior serving as the final classification layer. 

The continual learning scenario demonstrated in \cite{chip1, chip2} works as follows. The network is first trained using the seven first classes of the MNIST dataset in an offline training manner. Then, the remaining three last classes are learned by the sole STDP-driven adaptation of the output SNN layer (without access to the data of the first seven classes), while still retaining some knowledge of the past seven classes thanks to the Winner-Take-All mechanism. Using this continual learning setup, the authors in \cite{chip1, chip2} reach an overall accuracy of $93\%$. The same experiment is repeated with the CIFAR10 dataset, resulting in an overall accuracy of $82\%$ after continual learning. In essence, the authors in \cite{chip1, chip2} achieve a task-based continual learning setup comprised of two separate tasks to be learned.

An innovative aspect of the work in \cite{chip1, chip2} lies in the use of Phase Change Memory (PCM) material to store the synaptic weights of the SNN in the proposed circuit implementation. For this hardware perspective, the work in \cite{chip1, chip2} significantly departs from conventional \textit{digital} SNN-STDP processor design by exploring the use of highly novel and exploratory circuit elements such as PCMs. PCMs are a class of resistive memory that can store the synapse value as the value of their electrical resistance. A key aspect of PCM devices is that their electrical resistance value can be altered depending on the current and signals applied on the PCM terminals. This makes them particularly well-suited for local learning rules such as STDP which only depend on pre-synaptic and post-synaptic neural activity. Regarding energy consumption, their proposed system requires an average energy consumption of $\sim 0.1$ $\mu$J, which makes the proposed PCM-based SNN-STDP solution well-suited for extreme edge AI application. It is important to precise that other types of innovative nanodevices beyond PCMs are being currently studied by many teams for implementing synaptic weights, such as memristors \cite{memristormcmc}, ferroelectric memories \cite{ferro} and magnetoresistive memories \cite{magneto}.  

The work proposed in \cite{chip1, chip2} by I. Munoz-Martin, S. Bianchi \textit{et al.} constitutes a major stepping stone for the conception of energy-efficient edge AI devices via innovative circuit technologies that can continuously learn and personalize their underlying SNN model to new tasks and to changes in their working environment.

\begin{table}
\caption{\label{jlab2}Works in the field of continual STDP learning covered in Section \ref{clstdp}.}
\footnotesize
\centering
\begin{tabular}{@{}llll}
\br
Reference & CL Application & Architecture \\
\mr
D. I. Antonov \textit{et al.} \cite{antonov} & Task-Incremental &Multi-layer Convolutional SNN-STDP \\

A. Ororbia \cite{ororbia5} & Task-Incremental & Predictive Coding via multi-layer SNN-STDP \\

J. M. Allred \textit{et al.} \cite{alfredo} & Task-Incremental & Excitatory-Inhibitory with Dopaminergic feedback \\

A. Safa \textit{et al.} \cite{asafaslam2} & SLAM & Recurrent SNN-STDP \\

A. Safa \textit{et al.} \cite{asafapeople} & People Detection & Convolutional SNN-STDP\\

I. Munoz-Martin, \\
S. Bianchi \textit{et al.} \cite{chip1, chip2}
 & Task-Incremental & CNN feature map - SNN-STDP\\

\br
\end{tabular}\\

\end{table}

Table \ref{jlab2} briefly summarizes the works in the field of continual STDP learning that have been covered in this Section.

\section{Discussion and open future research questions}
\label{future}

The emerging number of works in neuromorphic continual learning covered in the Sections \ref{clhebbian} and \ref{clstdp} of this paper have clearly demonstrated the usefulness of bio-plausible neuromorphic architectures based on Sparse and Predictive Coding, as well as on Spiking Neural Networks with STDP learning for a wide range of different continual-learning scenarios and applications. These different applications ranged from task-incremental and class-incremental learning to drone navigation use cases such as Simultaneous Localization and Mapping (SLAM) and people detection on drones. In this Section, we aim at providing a discussion on the future perspectives for the field of neuromorphic continual learning so as to identify both the open research questions and the future application domains where we expect continual learning systems based on SNNs and other hardware-efficient neuromorphic technologies to bring great value.

\subsection{Neuromorphic task-incremental learning: open research questions}

In light of the various task-incremental works surveyed in this paper \cite{ororbia1, antonov, ororbia5, alfredo}, using both non-spiking Hebbian networks and SNN-STDP architectures, we can now formulate a number of open research questions for further pushing the boundaries of research in neuromorphic task-incremental learning systems:

\begin{itemize}
    \item A number of works based on Sparse and Predictive Coding using both non-spiking Hebbian learning and spiking SNN-STDP have been proposed, but none of them is built using a \textit{convolutional} architecture. Indeed, in the fully-connected cases, the input data and the labels are all fed as inputs to the predictive coding setups during learning. Then, during inference, only the data is fed as input and the labels are inferred by re-projection of the neural activity back into the input space. Even though this technique works for the fully-connected case, it is not fully clear if the same process would be effective in the convolutional case, since feeding both 2D images and 1D label vectors during learning could lead to a degraded performance, due to the signal dimensionality imbalance between the 2D image and 1D label case.

    \item Another interesting direction of research would be to explore how regularization-based continual learning techniques found in deep learning literature \cite{9349197} could be fused with neuromorphic continual learning techniques in order to combine the advantages of both fields. Hence, it would be interesting to study how to apply e.g., Elastic Weight Consolidation \cite{EWC} to the Predictive Coding Hebbian and SNN-STDP learning systems proposed in \cite{ororbia1, antonov, ororbia5, alfredo}.

    \item Finally, a number of works have pointed to possible links between both Hebbian and STDP-based learning, and the theory of Bayesian learning \cite{safaIWAI, maashebbianbayes}. indeed, Bayesian learning theory provides a principled way to study how continual learning can be achieved since the \textit{iterative Bayesian update} is insensitive to catastrophic forgetting. Iterative Bayesian update is performed by inferring a Posterior distribution $P(w|D)$ over the neural network weights $w$ given the data $D$, using the well-known Bayes' rule $P(w|D)=\frac{L(D|w) P(w)}{P(D)}$ (with $L(D|w)$ the likelihood distribution and $P(w)$ the prior) \cite{bayestutorial}. In fact, a number of regularization-based continual learning systems used in deep learning (such as the Elastic Weight Consolidation technique \cite{EWC}) can be directly linked to the iterative Bayes update, under a number of approximations and simplifications  \cite{EWC}. As future research, the same could be done for the neuromorphic case, by studying how Bayesian learning can be formulated in Hebbian learning and SNN-STDP systems.  
\end{itemize}

\subsection{Continual learning in neuromorphic robotics systems: open research questions}

Following the various robotics-related works covered in this paper \cite{asafaslam1, asafaslam2, asafapeople}, we can formulate the following open research questions for the field of neuromorphic continual learning research in robotics:

\begin{itemize}
    \item An emerging number of work exploring the use of both SC-based Hebbian learning \cite{asafaslam1} and SNN-STDP learning \cite{asafaslam2} for Simultaneous Localization and Mapping have been proposed. Thanks to their continual learning nature, the neural networks used in these works did not need to be trained offline but were rather learned online, as the drone was exploring the unknown environment. Even though useful, the demonstrations in \cite{asafaslam1, asafaslam2, asafapeople} have been provided in a single type of environment (an indoor warehouse). Hence, it might be interesting to investigate how these continual learning neuromorphic SLAM system behave when navigating from one environment to a totally different one (e.g., going from indoor to outdoor). It is expected that in this case, multiple Hebbian or SNN-STDP networks will be needed with each one dedicated to a particular environment. Then, a higher-order system detecting when the robot is passing from one environment to another can be used to switch between the different Hebbian and SNN-STDP ensembles (similar to the task selector network in \cite{ororbia1}).

    \item The work in \cite{asafapeople} has studied the continual learning of people detection from an event-based camera mounted on a drone. In \cite{asafapeople}, continual learning is done by using a stream of provided labels indicating where walking people are present on the output attention maps. In order to increase the usability of neuromorphic continual learning perception systems, it might be interesting to study how the use of precise labels in \cite{asafapeople} could be dropped and replaced with the use of e.g., binary reward signals indicating if the drone is coming too close to a human subject or not. Indeed, it is not expected that precise labels are available during the continual learning process for practical use cases. Hence, studying how binary alarm-type signals could be used instead of precise labels will help making continual learning neuromorphic perception systems such as \cite{asafapeople} better suited for practical use cases.

    \item In light of the emerging number of work about the use of Hebbian learning for agent control \cite{safaIWAI, ororbia3, ororbia4}, more research is needed for bridging the gap between continual learning neuromorphic systems and classical reinforcement learning techniques. Works such as \cite{safaIWAI} have shown the benefits of continual Hebbian learning in ANNs over conventional reinforcement learning setups in terms of a faster learning convergence and a reduced need for memory-expensive replay buffers. On the other hand, these demonstrations are still limited to rather simple reinforcement learning environments (such as the mountain car environment in \cite{safaIWAI}). Therefore, more research is needed for extending the applicability of neuromorphic agent control systems based on Hebbian learning rules to more complex problems.  
\end{itemize}
 
\subsection{Design of novel neuromorphic hardware for continual learning}

A major promise of neuromorphic computing hardware resides in its ultra-low-power aspect, as opposed to the more resource-expensive deep learning accelerator hardware \cite{odin}. The power-efficient aspect of neuromorphic system is even more crucial when it comes to providing \textit{learning} capability in extreme edge AI devices. Indeed, conventional deep learning systems rely on the \textit{back-propagation of error} algorithm during training which is known to be significantly more compute-expensive compared to the local learning rules such as Hebbian and STDP learning \cite{stdpalijon, khacefstdp}.

Since providing learning at the edge is believed to be a key enabler of the next generation of edge AI devices that can continuously personalize their models to their specific environments, a fast-growing body of work for the design of SNN hardware equipped with energy-efficient STDP learning has emerged. Still, most of the SNN-STDP hardware designs that have been proposed in literature are demonstrated in a \textit{continual learning} scenario but rather work in conventional offline training settings (with the exception of \cite{chip1, chip2} among others).

Hence, it is important to put a greather focus on the design of SNN-STDP hardware that can operate in continual learning settings. In order to achieve this aim, novel research directions could be adopted, such as:

\begin{itemize}
    \item Closely collaborating with the neuromorphic algorithm research community in order to implement successful continual learning SNN-STDP designs into dedicated integrated circuits, following an algorithm-hardware co-design approach.
    \item Studying how regularization-based continual learning methods can be simplified and implemented in dedicated hardware accelerators, since this class of method provide a good balance between performance and compute complexity in comparison to e.g., memory replay methods \cite{bookchapter}.
    \item Exploring how the currently-existing SNN-STDP hardware designs can still be utilized in a continual learning settings, and under which conditions with which impact on the performance.
\end{itemize}

\section{Conclusion}
\label{conclusion}

This paper has provided an overview of the rapidly-evolving field of neuromorphic continual learning using Sparse and Predictive Coding architectures, with Hebbian learning and its STDP variant in the spiking domain. First, this paper has motivated and described the background theory needed for the study of bio-plausible neural networks based on Sparse and Predictive Coding, learning via Hebbian and STDP rules. This paper has also provided background on continual learning by introducing the various methods used in literature to learn in non-stationary data streams. Then, after introducing the background theory, this paper has reviewed a number of recent work in the field of neuromorphic continual learning. The works surveyed in this paper have covered a wide range of different applications, from task-incremental and class-incremental learning to robotics, Simultaneous Localization and Mapping, and agent control. Finally, this paper has provided a future perspective on the open research questions for the field of neuromophic continual learning. It is hoped that this survey paper will be useful for the wider neuromorphic research community and will inspire future research in bio-plausible continual learning.

\end{document}